\documentclass[conference]{IEEEtran}
\usepackage{tabularx}

\usepackage{cite}
\usepackage{amsmath,amssymb,amsfonts}
\usepackage{algorithmic}
\usepackage{graphicx}
\usepackage{textcomp}
\usepackage{svg}
\usepackage[table]{xcolor}
\usepackage{booktabs}
\usepackage{tabularx}
\usepackage{array}
\usepackage{colortbl}
\usepackage{tikz}
\definecolor{TableNavy}{HTML}{173A5E}
\definecolor{TableGrid}{HTML}{B7C4CC}

\definecolor{FigBlue}{HTML}{2F80D0}
\definecolor{FigBlueDark}{HTML}{185B9D}
\definecolor{FigBlueLight}{HTML}{EAF5FD}
\definecolor{FigBlueStripe}{HTML}{EFF6FB}

\definecolor{FigOrange}{HTML}{F07A2A}
\definecolor{FigOrangeDark}{HTML}{A95A25}

\definecolor{FigGreen}{HTML}{16966A}
\definecolor{FigGreenDark}{HTML}{11724F}
\definecolor{FigGreenLight}{HTML}{EAF8F0}
\definecolor{FigGreenStripe}{HTML}{F0F8F3}

\definecolor{TableHead}{HTML}{E9F0F5}
\definecolor{TableText}{HTML}{1C3044}
\definecolor{TableLink}{HTML}{168C69}

\newcolumntype{L}[1]{>{\raggedright\arraybackslash}p{#1}}
\newcolumntype{C}[1]{>{\centering\arraybackslash}p{#1}}

\newcommand{\offlinebadge}{%
  \begingroup
  \setlength{\fboxsep}{1.0pt}%
  \colorbox{FigBlueLight}{%
    \textcolor{FigBlueDark}{\bfseries\strut offline}}%
  \endgroup
}

\newcommand{\onlinebadge}{%
  \begingroup
  \setlength{\fboxsep}{1.0pt}%
  \colorbox{FigGreenLight}{%
    \textcolor{FigGreenDark}{\bfseries\strut online}}%
  \endgroup
}

\newcommand{\benchlink}[1]{%
  \href{#1}{%
    \textcolor{TableLink}{%
      \raisebox{-0.08ex}{\small\faIcon{external-link-square-alt}}%
    }%
  }%
}
\usepackage[hidelinks]{hyperref}
\usepackage{fontawesome5}
\usepackage{longtable}
\usepackage{multirow}
\usepackage{pdflscape}
\usepackage{pifont}

\newcommand{\githublink}[1]{%
  \href{#1}{\raisebox{-0.05ex}{\faGithub}}%
}
\newcommand{\datasetlink}[1]{%
  \href{#1}{\raisebox{-0.05ex}{\faDatabase}}%
}
\newcommand{\projectlink}[1]{%
  \href{#1}{\raisebox{-0.05ex}{\faGlobe}}%
}

\usepackage{tikz}
\usetikzlibrary{arrows.meta,calc,positioning,fit,backgrounds}

\definecolor{taxRoot}{RGB}{45,91,163}
\definecolor{taxContext}{RGB}{205,205,245}
\definecolor{taxContextFill}{RGB}{247,247,255}
\definecolor{taxEvidence}{RGB}{89,188,198}
\definecolor{taxEvidenceFill}{RGB}{241,252,253}
\definecolor{taxTemporal}{RGB}{255,156,148}
\definecolor{taxTemporalFill}{RGB}{255,247,243}
\definecolor{taxMulti}{RGB}{183,237,112}
\definecolor{taxMultiFill}{RGB}{249,255,240}
\definecolor{taxTrunk}{RGB}{145,154,170}

\newcommand{\taxpaper}[2]{#1~\cite{#2}}

\def\BibTeX{{\rm B\kern-.05em{\sc i\kern-.025em b}\kern-.08em
    T\kern-.1667em\lower.7ex\hbox{E}\kern-.125emX}}
\makeatletter

\let\savedatmaketitle\@maketitle
\makeatother
\begin{document}

\title{Agentic Video Understanding: A Survey }

\author{
\IEEEauthorblockN{Xinyu Deng}
\IEEEauthorblockA{
\textit{The University of Western Australia} \\
ninadeng2023@gmail.com
}
\and
\IEEEauthorblockN{Siwen Luo}
\IEEEauthorblockA{
\textit{The University of Western Australia} \\
siwen.luo@uwa.edu.au
}
\and
\IEEEauthorblockN{Daochang Liu}
\IEEEauthorblockA{
\textit{The University of Western Australia} \\
daochang.liu@uwa.edu.au
}
}


\maketitle

\begin{abstract}
As large language models (LLMs) become capable of processing increasingly diverse modalities and longer temporal contexts, an emerging line of work is moving beyond fixed video-language inference toward agentic systems that actively decide what information to inspect, retain, verify, and act upon. This survey reviews video understanding agents: systems that use video as the primary information source and solve understanding tasks through adaptive state construction and action selection. We first formalize an agent loop for video understanding, then address a central question: why do agents matter for video understanding? To answer this, we organize the literature through a challenge-to-design taxonomy, linking context bottlenecks to hierarchical evidence memory, evidence sparsity to active evidence acquisition, temporal causality to state and process tracking, and multimodal ambiguity to role-specialized coordination. We further review state space paradigms, learning paradigms, supervision signals, benchmarks, and evaluation protocols. Finally, we identify open directions toward agentic-native temporal modeling and video-native agents.
\textbf{Project page:} \url{https://github.com/DXY0711/Awesome-Agentic-Video-Understanding}
\end{abstract}


\section{Introduction}

Video understanding has evolved from recognition-centered modeling to open-ended reasoning in long, multimodal, and interactive videos. Early \emph{video networks} treated video as a spatiotemporal signal and learned representations for action recognition, localization, captioning, and retrieval through temporal convolution, recurrent modeling, and space-time attention~\cite{tran2015learning,carreira2017quo,donahue2015long,bertasius2021space}. More recently, \emph{Video Large Language Models (Video LLMs)} have reframed video as input for language-based interaction, enabling models to answer questions, summarize events, localize moments, and reason about the information that video can provide, such as visual content, text captions, and timestamps~\cite{yang2023vid2seq,ren2024timechat,song2024moviechat,huang2024vtimellm}. This progression has greatly expanded the interface of video understanding, but conventional Video LLMs typically operate along a largely predetermined inference path: video evidence is constructed and provided to the model without allowing its intermediate reasoning state to control what should be inspected or revisited as reasoning~\cite{li2026lenswalk,zhang2026eva}.

The weakness of the pipeline becomes apparent when the relevant evidence is difficult to identify. Long videos often contain substantial temporal redundancy, while the answer may depend on a brief event, a distant earlier state, or a cue from speech, sound, text, or interaction. In streaming settings, the problem is even more difficult because future evidence may not yet be observable and an immediate answer could be premature. Under these conditions, video understanding is no longer only a problem of processing available evidence; it also becomes a problem of adaptively controlling where to look, what to store, which evidence to trust, whether more observation is needed, and when a response is justified.

\begin{figure}[t]
    \centering
    \includegraphics[width=\columnwidth]{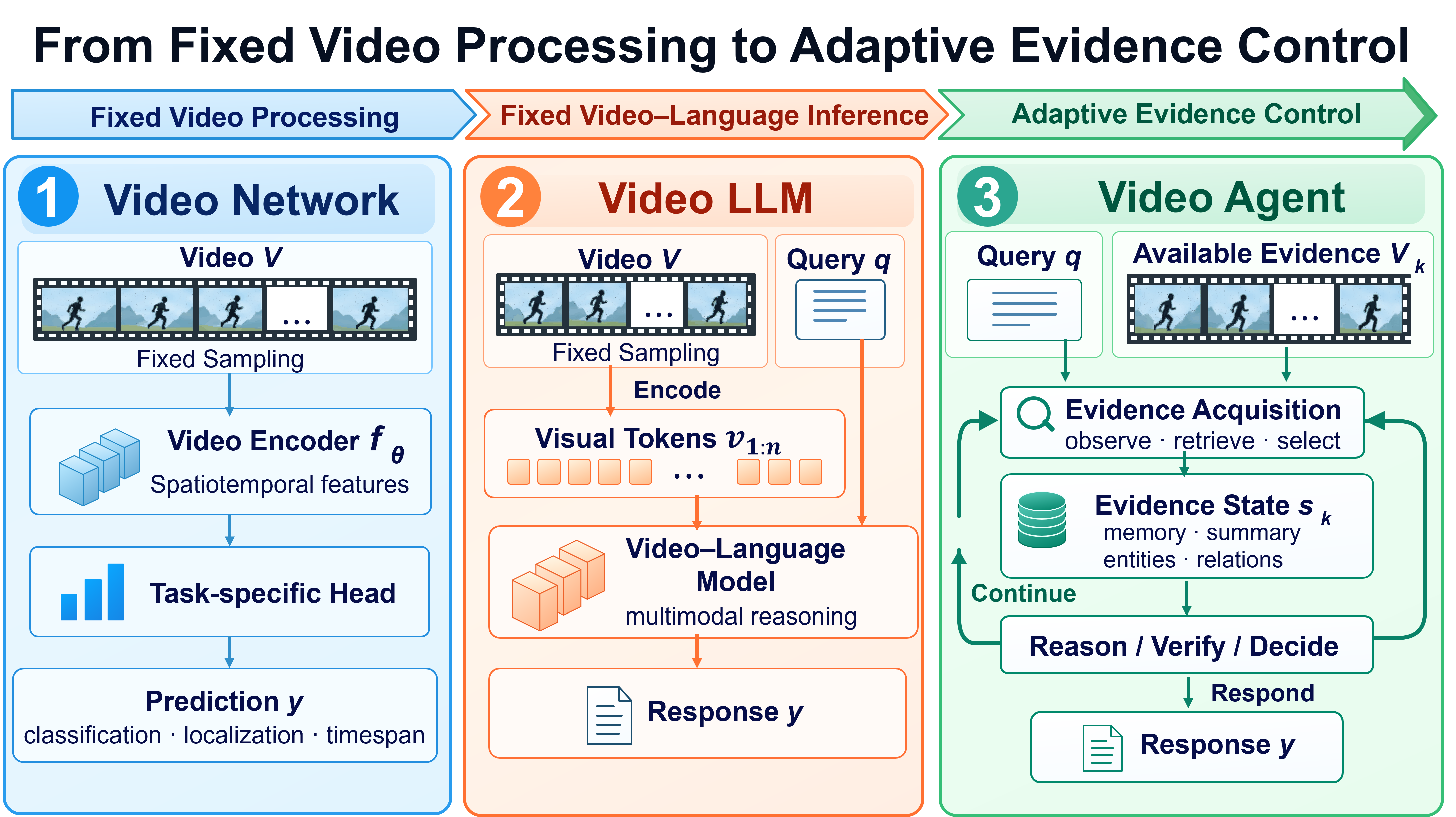}
    \caption{From Video Networks to Video Agents: The Shift from Passive Video Processing to Adaptive Evidence Control.}
    \label{fig:video_system_evolution}
\end{figure}

\emph{Video agents} emerge from this shift by introducing control into the video understanding loop. Rather than following a predetermined path from sampled frames to an answer, they instead couple reasoning with iterative evidence acquisition, allowing intermediate states to influence subsequent actions~\cite{wang2024videoagent,zhang2024omagent,ma2025drvideo,zhi2025videoagent2,zhang2025avila,liu2026thinking}. As illustrated in Fig.~\ref{fig:video_system_evolution}, this marks a shift from passive video processing to adaptive evidence control. A formal definition and the corresponding inclusion criteria are provided in Section~\ref{sec:scope}. 

Existing surveys focus on video-language understanding, video foundation models, and Video LLMs~\cite{nguyen2024video,madan2024foundation,tang2025video}. None of the surveys we identified systematically cover the emerging field of video agents. This survey addresses this gap by organizing agentic video understanding around the interaction between video-specific challenges and agentic control.
This survey follows the structure below.
\begin{enumerate}
    \item We define \emph{video understanding agents} as systems that solve video tasks through adaptive state construction and action selection, thereby distinguishing them from conventional Video LLMs with fixed inference pipelines.

    \item We develop a challenge-to-design taxonomy that explains why agentic mechanisms are needed for video understanding: context bottlenecks motivate hierarchical evidence memory, evidence sparsity motivates active evidence acquisition, temporal causality motivates state and process tracking, and multimodal ambiguity motivates role-specialized coordination.

    \item We introduce a state space view of agentic video understanding, tracing how video is represented as a bag of frames, a sequence of frames, a graph of entities, or an evolving world state.

    \item We further examine how video agent behaviors are specified or learned, covering inference-time control, supervised imitation, and reinforcement learning, together with the trajectory, grounding, and reward signals that support them.

    \item We review benchmarks, evaluations, and future directions for the field.

\end{enumerate}

\section{Scope and Definitions}
\label{sec:scope}
\subsection{Video Understanding Tasks}

We use \emph{video understanding} to refer to tasks in which video is the primary evidence source and successful prediction requires reasoning over information contained in the video, including its visual content, temporal structure, multimodal cues, or evolving state. This scope covers a broad range of video-conditioned tasks, including video question answering, captioning, temporal grounding, retrieval, summarization, and streaming video understanding.

\subsection{Video Understanding Agents}

To distinguish video understanding agents from fixed-inference Video
LLMs, we formalize video understanding as an adaptive decision process
over video evidence. Let
\[
V=\{x_{\tau}\}_{\tau=1}^{T}
\]
denote a temporally ordered video, where \(x_\tau\) denotes the multimodal observation available at time \(\tau\), including visual content and any video-derived auxiliary signals such as audio, text, and metadata. Let \(q\) denote the task instruction,
\(\mathcal{T}\) the available tools, and \(s_k\) the intermediate evidence
state at decision step \(k\). We use \(\bar V_k\) to denote the video
evidence available at that step: \(\bar V_k=V\) for offline video, while
in streaming settings it contains only the observed prefix.

We define a \emph{video understanding agent} as a system that uses
video as its primary source and solves the task through
adaptive evidence-state construction and action selection:
\begin{align}
a_k &\sim \pi_{\theta}(a\mid q,s_{k-1},\bar V_k,\mathcal{T}),\\
o_k &= \mathcal{O}(\bar V_k,a_k,\mathcal{T}),\\
s_k &= \mathcal{U}(s_{k-1},o_k,a_k,q).
\end{align}
Here, an action may change what evidence is accessed, how the state is updated, whether evidence is verified, or when the agent stops to produce an output. The resulting observation \(o_k\) is incorporated into the
updated evidence state \(s_k\). If
\(K=\min\{k:a_k=\textsc{Stop}\}\), the final output is
\[
y=\mathcal{D}_{\theta}(q,s_K).
\]

A system qualifies as a video understanding agent when it maintains an intermediate evidence state and adaptively selects at least one action that affects subsequent evidence access, state evolution, tool use, interaction, or termination. Fixed-sampling, one-pass Video LLMs are therefore excluded.

\subsection{Survey Corpus and Inclusion Criteria}
We construct the survey corpus primarily from recent papers in major machine learning and computer vision venues, including ICML, ICLR, NeurIPS, CVPR, ICCV, and ECCV, along with highly relevant recent preprints that introduce agentic video-understanding systems not yet represented in archival venues.  We focus on papers published or released between 2024 and 2026, as most of the literature fitting our definition emerged during this period, while also including earlier foundational benchmarks, datasets, and video-understanding methods when necessary to provide task or evaluation context.

A work is included in the core corpus when it satisfies three criteria:
\begin{itemize}
    \item video serves as a primary input source; 
    \item the task falls within the scope of video understanding defined above; 
    \item the method exhibits agentic control as defined in the Video Understanding Agents subsection. 
\end{itemize}
Under these criteria, the survey covers offline and online settings, short and long videos, single-video and multi-video tasks, and emerging interactive or streaming agentic tasks.

Adjacent areas such as embodied or egocentric agents that learn from video demonstrations~\cite{fan2025embodied,han2025roomtour3d,chen2025egoagent}, computer-use agents trained from screen recordings~\cite{lu2025videoagenttrek,jang2025scalable,wang2025mobile}, and video generation or editing agents~\cite{wang2024lave,tu2026spagent} are considered only when they clarify mechanisms relevant to video understanding, rather than as primary targets of the survey.

\section{Challenge-to-Design Taxonomy}
The central claim of our taxonomy is that video agents should be organized not merely by the mechanisms they use, but by the video-specific challenge that makes an agentic mechanism matter. Instead of asking whether a system contains memory, tools, state representations, or multiple agents, we ask what bottleneck in video understanding these designs address. General-purpose agents also use memory, tools, state, and collaboration, but video gives these designs sharper roles: memory retains evidence under long temporal and token budgets, search acquires sparse query-relevant evidence, state tracking models temporal change and causal continuity, and collaboration coordinates heterogeneous modalities and perceptual roles. Because a single method may combine modules that address different bottlenecks, we assign each paper to the primary challenge foregrounded in its problem formulation and directly targeted by its central contribution.

\begin{figure*}[!t]
\centering
\resizebox{\textwidth}{!}{%
\begin{tikzpicture}[
    line cap=round,
    line join=round,
    root/.style={
        draw=taxRoot,
        fill=taxRoot!4,
        line width=1.05pt,
        rounded corners=4pt,
        minimum width=4.65cm,
        minimum height=0.68cm,
        inner xsep=6pt,
        font=\bfseries\fontsize{9.2}{10.4}\selectfont,
        align=center
    },
    challenge/.style={
        rounded corners=3pt,
        minimum width=4.2cm,
        minimum height=0.65cm,
        line width=0.95pt,
        font=\bfseries\fontsize{8.4}{9.5}\selectfont,
        align=center,
        inner xsep=6pt,
        inner ysep=4pt
    },
    papers/.style={
        rounded corners=11pt,
        line width=1.15pt,
        text width=13.85cm,
        align=left,
        inner xsep=9pt,
        inner ysep=8pt,
        font=\fontsize{8}{9}\selectfont,
        execute at begin node=\raggedright
    },
    branch/.style={line width=0.85pt},
    arrow/.style={-{Stealth[length=2.2mm,width=1.5mm]},line width=0.85pt}
]

\node[root, rotate=90] (root) at (0,-0.65) {Challenge-to-Design Taxonomy};
\coordinate (trunk) at (1.25,-0.65);
\draw[branch,draw=taxRoot] (0.36,-0.65) -- (trunk);
\draw[branch,draw=taxTrunk!65] (1.25,4.95) -- (1.25,-5.15);

\node[challenge,draw=taxContext,fill=taxContextFill] (context) at (4.25,4.95) {Context Bottleneck};
\node[challenge,draw=taxEvidence,fill=taxEvidenceFill] (evidence) at (4.25,2.10) {Evidence Sparsity};
\node[challenge,draw=taxTemporal,fill=taxTemporalFill] (temporal) at (4.25,-1.55) {Temporal Causality};
\node[challenge,draw=taxMulti,fill=taxMultiFill] (multi) at (4.25,-5.15) {Multimodal Ambiguity};

\node[papers,draw=taxContext,fill=taxContextFill,anchor=west] (contextpapers) at (7.10,4.95) {
\taxpaper{DrVideo}{ma2025drvideo},
\taxpaper{HAVEN}{yin2026hierarchical},
\taxpaper{VideoARM}{yin2026videoarm},
\taxpaper{AVI}{gao2025agentic},
\taxpaper{Flash-VStream}{zhang2025flash},
\taxpaper{StreamChat}{xiong2025streaming},
\taxpaper{ProVideLLM}{chatterjee2025memory},
\taxpaper{Video-RAG}{luo2026video},
\taxpaper{StreamMeCo}{wang2026streammeco},
\taxpaper{AdaVideoRAG}{zhang2026adavideorag},
\taxpaper{Mr. Video}{pang2025mr},
\taxpaper{VideoAgent (Fan et al.)}{fan2024videoagent},
\taxpaper{R3-Streaming}{liu2026efficient},
\taxpaper{VideoLLaMB}{wang2025videollamb},
\taxpaper{StreamRAG}{xie2026streamrag},
\taxpaper{VideoLucy}{zuo2026videolucy},
\taxpaper{WorldMM}{yeo2026worldmm},
\taxpaper{VideoStreaming}{qian2024streaming},
\taxpaper{ReKV}{di2025streaming},
\taxpaper{M3-Agent}{long2025seeing},
\taxpaper{OmAgent}{zhang2024omagent},
\taxpaper{SALOVA}{kim2025salova},
\taxpaper{SlowFocus}{nie2024slowfocus},
\taxpaper{LVAgent}{lvagent2025},
\taxpaper{EGAgent}{rege2026agentic},
\taxpaper{OmniAgent}{xing2026native}, and
\taxpaper{Symphony}{yan2026symphony}.
};

\node[papers,draw=taxEvidence,fill=taxEvidenceFill,anchor=west] (evidencepapers) at (7.10,2.10) {
\taxpaper{A4VL}{a4vl2026},
\taxpaper{Commonsense Video QA}{liu2025commonsense},
\taxpaper{DVD}{zhang2026deep},
\taxpaper{AoTD}{shi2025enhancing},
\taxpaper{EVA}{zhang2026eva},
\taxpaper{LensWalk}{li2026lenswalk},
\taxpaper{ReAgent-V}{reagentv2025},
\taxpaper{SAGE}{jain2026sage},
\taxpaper{Thinking with Videos}{zhang2026thinking},
\taxpaper{VCA}{yang2025vca},
\taxpaper{VideoAgent (Wang et al.)}{wang2024videoagent},
\taxpaper{VideoAgent2}{zhi2025videoagent2},
\taxpaper{VideoChat-A1}{wang2026videochat},
\taxpaper{VideoExplorer}{yuan2025videoexplorer},
\taxpaper{VideoSeek}{lin2026videoseek},
\taxpaper{APPO}{du2026appo},
\taxpaper{LongVideo-R1}{qiu2026longvideo},
\taxpaper{MSR-ViR}{song2025modularized},
\taxpaper{ReViSe}{xu2026towards},
\taxpaper{LongVT}{yang2026longvt},
\taxpaper{Select Less, Reason More}{li2026select},
\taxpaper{FrameThinker}{he2025framethinker},
\taxpaper{Video-MTR}{xie2025video},
\taxpaper{VideoBrain}{zou2026videobrain},
\taxpaper{VideoSEAL}{qiu2026videoseal},
\taxpaper{Refer-Agent}{jiang2026referagent},
\taxpaper{VideoChat-M1}{chen2026videochat}, and
\taxpaper{VideoHV-Agent}{wang2026think}.
};

\node[papers,draw=taxTemporal,fill=taxTemporalFill,anchor=west] (temporalpapers) at (7.10,-1.55) {
\taxpaper{AVT}{yang2024agent},
\taxpaper{AViLA}{zhang2025avila},
\taxpaper{DoraemonGPT}{yang2024doraemongpt},
\taxpaper{EgoAgent}{chen2025egoagent},
\taxpaper{Embodied VideoAgent}{fan2025embodied},
\taxpaper{GraphVideoAgent}{chu2025graphvideoagent},
\taxpaper{Streaming Harness}{yao2026harnessing},
\taxpaper{Mobile-Agent-V}{wang2025mobile},
\taxpaper{PANDA}{yang2026panda},
\taxpaper{ReWatch-R1}{zhang2025rewatch},
\taxpaper{RoomTour3D}{han2025roomtour3d},
\taxpaper{MONDAY}{jang2025scalable},
\taxpaper{SVAgent}{yang2026svagent},
\taxpaper{ThinkStream}{liu2026thinking},
\taxpaper{VideoAgentTrek}{lu2025videoagenttrek},
\taxpaper{VideoMind}{liu2025videomind},
\taxpaper{VTimeCoT}{zhang2025vtimecot},
\taxpaper{Takusen}{tang2026asynchronous},
\taxpaper{Eyes Wide Open}{yu2026eyes},
\taxpaper{LION-FS}{li2025lion},
\taxpaper{SEASON}{wu2026season},
\taxpaper{StreamAgent}{yang2025streamagent},
\taxpaper{StreamBridge}{wang2026streambridge},
\taxpaper{Time-R1}{wang2026time},
\taxpaper{Video-R1}{feng2026video},
\taxpaper{VITED}{lu2025vited},
\taxpaper{When Thinking Drifts}{luo2026thinking},
\taxpaper{Video-of-Thought}{pmlr-v235-fei24a},
\taxpaper{StreamReady}{azad2026streamready},
\taxpaper{Proact-VL}{yan2026proact},
\taxpaper{Thinking-QwenVL}{zhang2026progressive}, and
\taxpaper{AgenticVS}{guo2026agentic}.
};

\node[papers,draw=taxMulti,fill=taxMultiFill,anchor=west] (multipapers) at (7.10,-5.15) {
\taxpaper{MAViD}{chen2025multimodal},
\taxpaper{IXC2.5-OL}{zhang2024internlm},
\taxpaper{MAGNET}{chowdhury2026magnet},
\taxpaper{SciEducator}{xu2026scieducator},
\taxpaper{V-Agent}{park2025v},
\taxpaper{ViSpeak}{fu2025vispeak}, and
\taxpaper{G2F-RAG}{yang2026graph}.
};

\draw[arrow,draw=taxContext] (1.25,4.95) -- (context.west);
\draw[arrow,draw=taxEvidence] (1.25,2.10) -- (evidence.west);
\draw[arrow,draw=taxTemporal] (1.25,-1.55) -- (temporal.west);
\draw[arrow,draw=taxMulti] (1.25,-5.15) -- (multi.west);

\draw[branch,draw=taxContext] (context.east) -- (contextpapers.west);
\draw[branch,draw=taxEvidence] (evidence.east) -- (evidencepapers.west);
\draw[branch,draw=taxTemporal] (temporal.east) -- (temporalpapers.west);
\draw[branch,draw=taxMulti] (multi.east) -- (multipapers.west);

\end{tikzpicture}%
}
\caption{Challenge-to-design taxonomy of video understanding agents. Each branch links a video-specific bottleneck to representative agentic systems that address it.}
\label{fig:challenge_taxonomy}
\end{figure*}
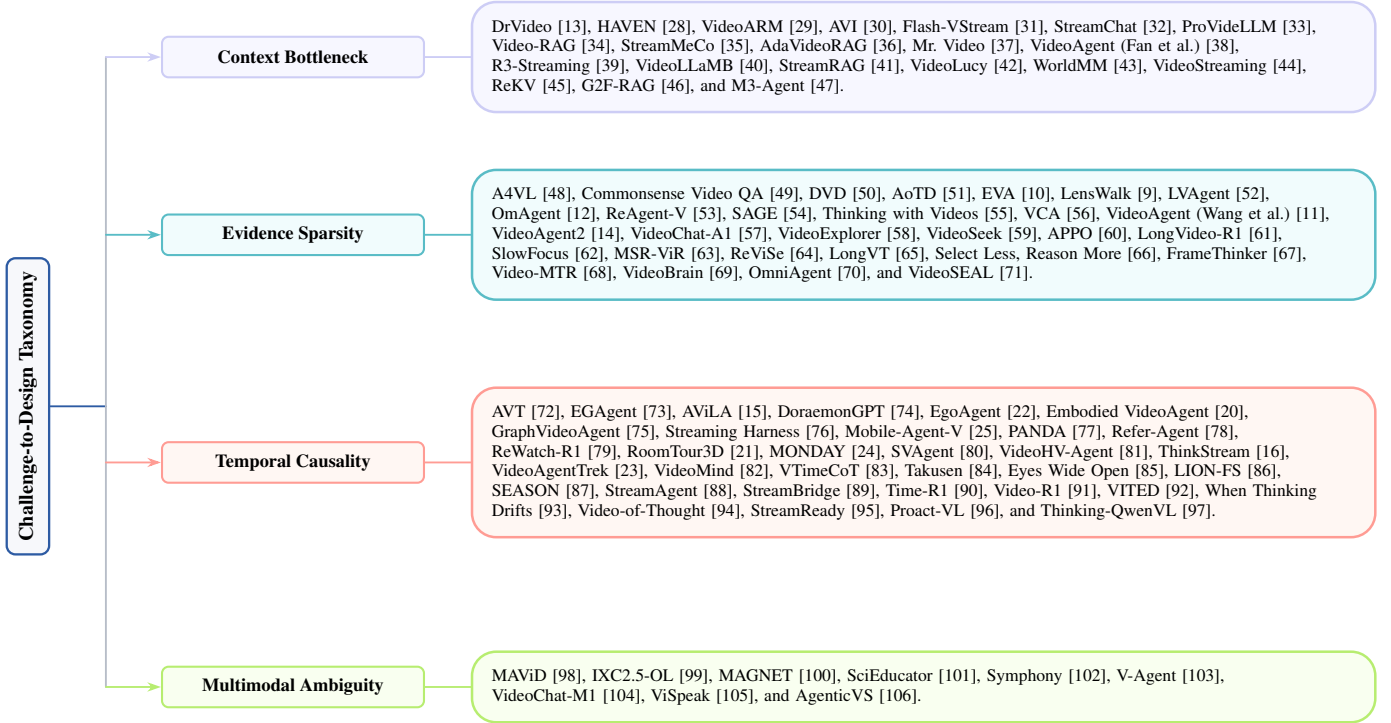

The complete challenge-to-design taxonomy is provided in
Fig.~\ref{fig:challenge_taxonomy}.

\label{app:challenge_to_design_tree}

\subsection{Context Bottleneck: Why Video Agents Need Hierarchical Memory}
The context bottleneck is a central challenge in video understanding and becomes especially severe in long-form, egocentric, multi-source, or streaming settings. Conventional Video LLMs operate under finite context and computational budgets, which often motivate sparse frame sampling, visual-token reduction, or compression of long visual histories~\cite{shen2024longvu,li2024llama,song2024moviechat}. Such reductions are inherently lossy: long videos contain substantial redundant content, while task-relevant evidence may depend on fine-grained visual details or events occurring at different temporal scales that can be discarded during abstraction. Consequently, representing a long video with a single flat or fixed-scale context can make it difficult to preserve both global structure and fine-grained evidence.

Hierarchical evidence memory addresses this bottleneck by making video evidence addressable at multiple temporal and semantic levels. WorldMM~\cite{yeo2026worldmm} separates episodic, semantic, and visual memories, allowing a retrieval agent to choose the memory source and scale that fit the query. VideoARM~\cite{yin2026videoarm} uses an observe-think-act-memorize loop to build hierarchical multimodal memory rather than repeatedly reprocessing the same evidence. HAVEN~\cite{yin2026hierarchical} adds audiovisual entity cohesion, preserving links between people, objects, sounds, and temporal segments. AVI~\cite{gao2025agentic} builds an entity-graph video knowledge base with retrieve-perceive-review.

For video agents, provenance and temporal granularity are particularly important because compressed memory may later serve as evidence for reasoning, retrieval, or decision making. Memory representations should therefore preserve sufficient links to their underlying video evidence, such as relevant frames or temporal intervals and the associated modalities or entities. Otherwise, abstraction and compression may introduce factual or temporal distortions that propagate into subsequent reasoning. Effective video memory thus involves a trade-off: it should be compact enough for efficient reasoning while retaining sufficient grounding to recover or verify the evidence when needed.

This trade-off becomes more pronounced under streaming and long-horizon input regimes, where retaining the full visual history is often impractical. Flash-VStream~\cite{zhang2025flash}, StreamChat~\cite{xiong2025streaming}, and ProVideLLM~\cite{chatterjee2025memory} use cache or memory structures to avoid repeatedly processing the full visual history, while StreamMeCo~\cite{wang2026streammeco} studies how to compress long-term agent memory without losing retrieval quality. Their contribution is to make memory operations explicit under severe compression and retrieval constraints.

\subsection{Evidence Sparsity: Why Video Agents Need Active Evidence Acquisition}
Many long-form video understanding tasks exhibit a needle-in-a-haystack structure, where task-relevant evidence may occupy only a small portion of the full video~\cite{ye2025re}. Fixed frame sampling is vulnerable when critical visual evidence falls outside the sampled observations~\cite{hu2025m,li2026lenswalk}. In this setting, a complementary challenge to memory is deciding what additional evidence to inspect next. Agentic approaches address this challenge by turning evidence acquisition into an adaptive process: the agent observes partial evidence, reasons about what information is still missing, selects a subsequent observation or tool action, and iterates as needed

Early video agents make the next observation decision explicit. VideoAgent~\cite{wang2024videoagent} uses an LLM controller to plan subgoals and retrieve visual evidence relevant to the task. VCA~\cite{yang2025vca} explores candidate video segments according to curiosity signals, while VideoAgent2~\cite{zhi2025videoagent2} uses uncertainty-aware reasoning to determine when additional evidence retrieval is needed and to adapt its retrieval plan accordingly. VideoSeek~\cite{lin2026videoseek} and LensWalk~\cite{li2026lenswalk} extend this loop by letting the agent choose among tools, temporal scopes, and sampling densities; EVA~\cite{zhang2026eva} makes the seek, plan, and reflect process trainable. Flow4Agent~\cite{liu2025flow4agent} is a related efficiency case, using motion priors to suppress redundant temporal and spatial information before reasoning.

This acquisition problem also spans tools and temporal scales. DVD~\cite{zhang2026deep} treats long video as a searchable database with tool calls at different levels of granularity, while VideoExplorer~\cite{yuan2025videoexplorer} decomposes questions, grounds relevant moments, and performs perception guided by the task.

Online settings extend this decision beyond selecting evidence from the observed video to deciding whether further observation is needed before a response is justified.

\subsection{Temporal Causality: Why Video Agents Need State Tracking}
Video is not only a collection of visual observations but an ordered record of change. Many questions cannot be answered by retrieving a single salient frame or segment. They ask what happened before another event, whether an object changed state, why a person acted, how a scene evolved, or which entity persisted across cuts and viewpoints. Frame- or segment-level evidence can be insufficient when the answer depends on state transitions or relations distributed across time. This challenge is distinct from memory and active acquisition. Memory concerns whether past evidence remains accessible, and active acquisition concerns where additional evidence should be found, whereas state tracking concerns how observed evidence is linked into evolving entities, relations, actions, and event chains.

State tracking makes temporal structure explicit. It can appear as an entity timeline, a scene graph that is updated over time, a storyline, a hypothesis ledger, or a memory in which each entry records what changed and when. SVAgent~\cite{yang2026svagent} organizes video understanding around a storyline that guides later reasoning and GraphVideoAgent~\cite{chu2025graphvideoagent} represents long videos through entity-relation graphs that preserve objects, events, and their temporal relations.

A memory may record that a person enters a kitchen; state tracking additionally represents whether the person later leaves, what they carry, and which subsequent events depend on that change. This distinction becomes especially important in long-horizon and online settings, where the agent must preserve and update such dependencies as new observations arrive. AViLA~\cite{zhang2025avila} addresses asynchronous query and evidence arrival, requiring the agent to distinguish observed evidence from information that is not yet available. ThinkStream~\cite{liu2026thinking} and StreamEQA~\cite{wang2026streameqa} examine related forms of incremental temporal-state reasoning, where an answer depends on past evidence, the current state, or the appropriate time to respond.

\subsection{Multimodal Ambiguity: Why Video Agents Need Role-Specialized Coordination}
Video evidence is multimodal and often ambiguous. For instance, the visual stream may show an action without revealing intent; speech may explain an event that is off camera; audio may identify a sound source before it appears visually; Optical Character Recognition (OCR) may disambiguate a location; and motion may contradict a static frame caption. A single model can integrate these signals, but multimodal models often exhibit modality imbalance or dominance, causing predictions to rely disproportionately on one modality and making them vulnerable when evidence from different modalities conflicts~\cite{sun2024video,zhang2025robust}.

One agentic design response is role-specialized multimodal coordination, where distinct perceptual or reasoning roles separate and reconcile heterogeneous evidence across modalities. Such specialization may be implemented through either multi-agent collaboration or dedicated modules.

Module-specialized multimodal systems are relevant because they separate functions that a monolithic model may blur. OmniLive~\cite{zhang2024internlm} separates perception, memory, and reasoning for continuous audiovisual interaction. MAViD~\cite{chen2025multimodal} uses video-audio multitask modeling and state-machine scheduling to
route and integrate heterogeneous signals, while ViSpeak~\cite{fu2025vispeak} studies visually expressed instructions such as gesture-based wake-up, interruption, and termination.

Multimodal ambiguity may also arise when evidence must be reconciled across representations or sources. G2F-RAG~\cite{yang2026graph} addresses competition between retrieved textual knowledge and visual video evidence by rendering structured knowledge into the visual representation space. MAGNET~\cite{chowdhury2026magnet} uses multi-agent reasoning to retrieve and integrate sparse audio-visual evidence distributed across multiple videos.

Overall, this dimension concerns how specialized roles or modules reconcile heterogeneous video evidence across modalities, representations, and sources. Beyond final accuracy, evaluation should assess whether such coordination improves evidence coverage, cross-modal correction, and groundedness.

\section{Paradigm: State Space}
Having linked video-specific challenges to agentic design choices, we
next examine the operative video states on which agents reason and act.
We organize existing methods into four dominant paradigms:
\emph{Video as a Bag of Frames}, \emph{Video as a Sequence of Frames},
\emph{Video as a Graph of Entities}, and \emph{Video as an Evolving
World State}. These paradigms differ in whether video evidence is
organized as discrete visual units, temporally related observations,
persistent relational structures, or causally updated states as the
video unfolds.

\begin{figure}[t]
    \centering
    \includegraphics[width=\columnwidth]{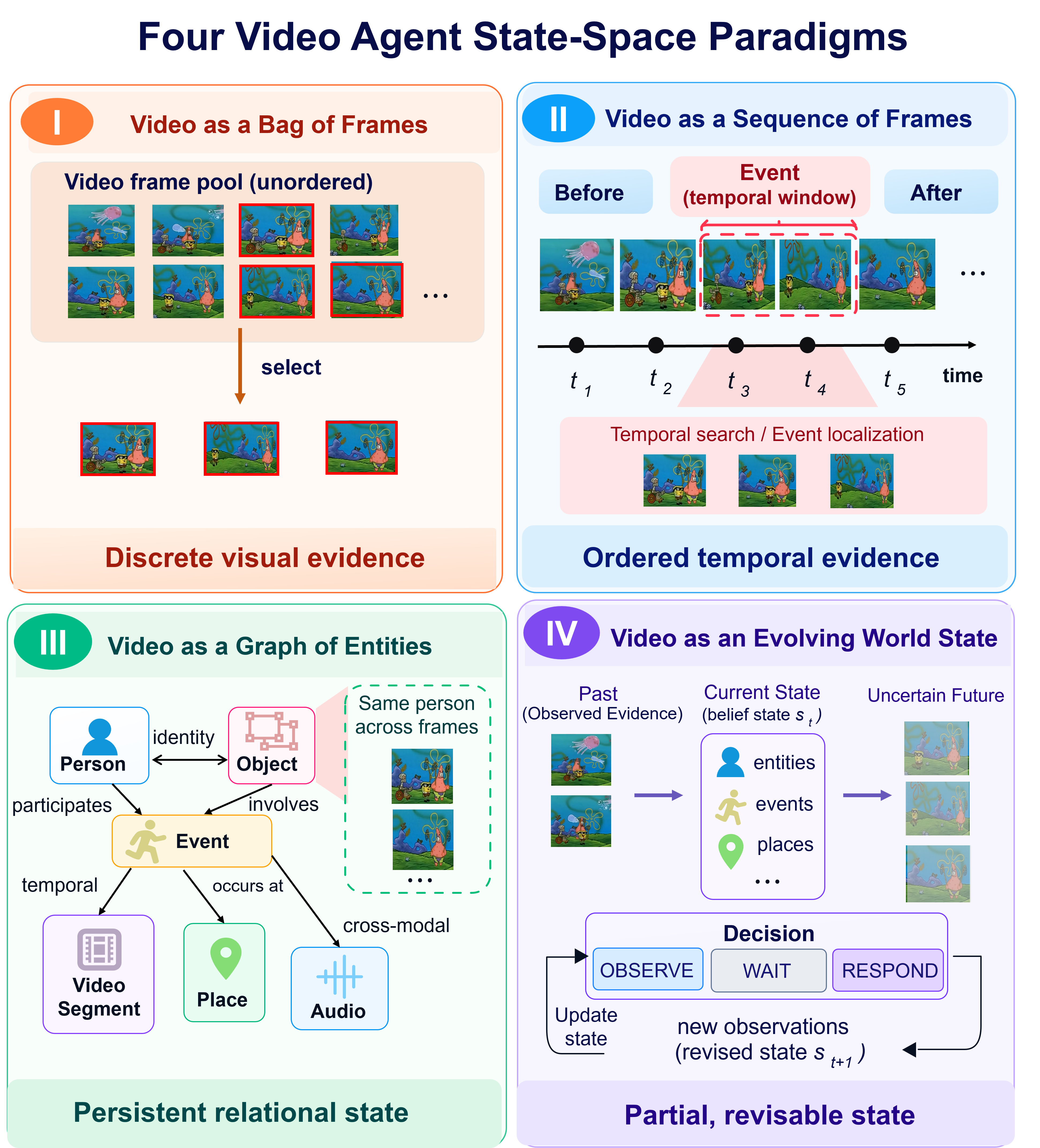}
    \caption{Four Video Agent State Space Paradigms.}
    \label{fig:State-Space Paradigms}
\end{figure}

Fig.~\ref{fig:State-Space Paradigms} illustrates these distinct state space paradigms and their progression in representational structure.

\subsection{Paradigm I: Video as a Bag of Frames}
In this paradigm, a video is not primarily represented as a continuous temporal process or as persistent long-term memory, but is instead abstracted into discrete visual evidence units, such as frames, keyframes, clips, shots, or candidate video segments. Methods in this paradigm may retain local temporal cues or basic ordering information, but they do not explicitly organize the evidence around temporal transitions, event evolution, or long-term state dependencies. Once a frame, clip, or segment is selected, it mainly functions as evidence for recognition, grounding, or answer generation. The resulting video state is therefore closer to a collection of query-relevant visual observations than to an explicit model of how events or states evolve over time.

Different methods instantiate this discrete evidence state at different granularities. EVA~\cite{zhang2026eva} adaptively acquires visual observations at selected temporal ranges and resolutions through iterative planning and perception, without constructing an explicit persistent temporal state over event transitions. Refer-Agent~\cite{jiang2026referagent} iteratively selects query-relevant keyframes, grounds target objects on a selected keyframe, and uses the resulting localization to initialize mask propagation across the video.

At a coarser granularity, A4VL~\cite{a4vl2026} partitions long videos into event-driven candidate blocks and iteratively selects query-relevant blocks and frames through clue-based alignment. LVAgent~\cite{lvagent2025} retrieves query-related video chunks and frames in each round, using them as evidence for subsequent multi-agent reasoning and collaboration.

Overall, Video as a Bag of Frames can offer a compact, efficient state-space abstraction compatible with existing VLMs and directly grounded in visual evidence. However, without explicit modeling of temporal order, entity persistence, or state transitions, it remains limited for reasoning about how events unfold and the video world changes over time.

\subsection{Paradigm II: Video as a Sequence of Frames}

Paradigm II represents video as an ordered sequence of observations. Compared with the bag-of-frames abstraction, the video state depends not only on which frames, clips, or segments are retained, but also on their positions and relations along the timeline. Methods in this paradigm make temporal structure explicit, so they interpret a segment through its position in an unfolding process rather than as isolated evidence.

Many long-video agents instantiate this paradigm through temporally organized evidence spaces. VideoAgent \cite{wang2024videoagent} and VideoAgent2 \cite{zhi2025videoagent2} have search and re-observation policies that operate over temporally localized segments and preserve temporal positions during evidence gathering. VCA \cite{yang2025vca} organizes long videos into hierarchical temporal segments, while LensWalk \cite{li2026lenswalk} treats observation as a multi-scale path through the video. VideoChat-A1 \cite{wang2026videochat} further decomposes videos into shot and subshot chains, allowing reasoning to follow their temporal organization. In these methods, the relations among selected segments, rather than segment relevance alone, contribute to the operative video state.

This sequential abstraction is also central to procedure and trajectory modeling. Mobile-Agent-V \cite{wang2025mobile} and VideoAgentTrek \cite{lu2025videoagenttrek} model screen recordings as ordered GUI state transitions. RoomTour3D \cite{han2025roomtour3d} represents room-tour videos as spatiotemporal trajectories for embodied navigation. AVT~\cite{yang2024agent} similarly evaluates clips within a coherent sequence, such that their meaning depends partly on preceding and subsequent observations.

In conclusion, these methods structure video understanding around temporal order and relations among observations, while the next paradigm shifts the state representation toward persistent entity-centric relations.

\subsection{Paradigm III: Video as a Graph of Entities}

Paradigm III organizes video as a relational state rather than only as an ordered sequence of observations. Here, an ``\textbf{entity}'' denotes any reusable state unit that can persist, recur, or support later reasoning across the video. Likewise, a ``\textbf{graph}'' need not be implemented as a formal graph data structure, but it must make associations across frames, segments, modalities, or reasoning steps explicit. These associations may represent identity, co-reference, temporal or spatial relations, multimodal alignment, or evidential support. Unlike Paradigm II, temporal order may still be preserved, but it is no longer the sole structure organizing video evidence.

A first line of work builds structured evidence archives from long videos. DrVideo \cite{ma2025drvideo} converts videos into document-style memory, where segment descriptions can be retrieved and supplemented with visual details. OmAgent \cite{zhang2024omagent} constructs a multimodal memory from scene captions, automatic speech recognition (ASR) transcripts, face recognition, timestamps, and vector databases. Video-RAG \cite{luo2026video} and AdaVideoRAG \cite{zhang2026adavideorag} further organize OCR, ASR, object texts, visual embeddings, and knowledge-graph-like records into retrievable evidence states. Mr. Video \cite{pang2025mr} follows a MapReduce formulation, aggregating scene-level captions while resolving references to persons and objects.

A second line makes entities and relations more central. VideoAgent \cite{fan2024videoagent} maintains temporal and object memory for long-video understanding. GraphVideoAgent \cite{chu2025graphvideoagent} explicitly represents long videos with entity-relation graphs. EGAgent \cite{rege2026agentic} extends this idea to days- or weeks-long egocentric videos through entity scene graphs over people, places, objects, and events. HAVEN \cite{yin2026hierarchical} uses audiovisual entity cohesion to connect visual, audio, and textual evidence around persistent entities. WorldMM \cite{yeo2026worldmm} further separates episodic, semantic, and visual memory, producing a multi-scale state that links events, long-term relations, and visual details.
Relational video states may also be constructed dynamically around the query. VideoARM \cite{yin2026videoarm} organizes video intervals, multimodal clues, and tool output in hierarchical memory.

In summary, this paradigm supports long-range association, multi-hop retrieval, entity persistence, and multimodal evidence fusion. Its main risk is representational distortion: if captions omit details, graphs link entities incorrectly, or memory states compress away relevant evidence, subsequent reasoning may be structurally organized but factually misgrounded.

\subsection{Paradigm IV: Video as an Evolving World State}

Paradigm IV represents video as a causally updated state of an unfolding situation. Unlike Paradigm III, whose defining structure lies in explicit associations among evidence units, Paradigm IV is defined by causal state maintenance: the representation must remain valid as new observations arrive while future frames remain unavailable. The state is therefore time-indexed, partial, and incrementally revisable. It captures what has been observed, the current interpretation of the video, and how incoming evidence changes that interpretation; it may also track uncertainty, missing evidence, or response readiness.

This distinction is most visible in streaming and interactive settings. AViLA \cite{zhang2025avila} addresses the asynchronous arrival of user queries and supporting visual evidence, requiring the system to distinguish observed from not-yet-observed information. ThinkStream \cite{liu2026thinking} frames streaming understanding as an incremental Watch--Think--Speak process, in which each incoming chunk may revise the current state before the system produces a response. R3-Streaming \cite{liu2026efficient} further connects state updates with readiness estimation, forgetting, and adaptive fast-slow routing. Streaming Harness \cite{yao2026harnessing}, ViSpeak \cite{fu2025vispeak}, and OmniLive \cite{zhang2024internlm} extend this setting to open-ended interaction and long-term audiovisual dialogue, where the system must jointly maintain evolving video and conversational context.

Other works emphasize prediction, risk, and embodied scene updates. StreamAgent \cite{yang2025streamagent} maintains a state that supports anticipation of possible future events rather than only summarizing past observations. Embodied VideoAgent \cite{fan2025embodied} updates object locations and scene states from egocentric video and embodied sensing. PANDA \cite{yang2026panda} treats surveillance video as an evolving risk context that combines scene information, anomaly rules, and uncertain observations. EgoAgent \cite{chen2025egoagent} represents egocentric interaction through interleaved state-action tokens, linking observation, prediction, and action. Taken together, this paradigm is defined not by online memory alone, but by whether the video representation remains causally valid and continuously revisable as the video unfolds.
%
\definecolor{methodrowblue}{HTML}{F7F9FB}
\definecolor{tablegrid}{HTML}{B8C2C9}
\definecolor{metablue}{HTML}{185B9D}
\definecolor{metabluelight}{HTML}{EEF5FA}
\definecolor{learningorange}{HTML}{C65D1E}
\definecolor{learningorangelight}{HTML}{FDF3EA}
\definecolor{datagreen}{HTML}{11724F}
\definecolor{datagreenlight}{HTML}{EDF7F1}
\definecolor{inactivegray}{HTML}{D7DCE1}

\definecolor{p1bg}{HTML}{E5F1FC}
\definecolor{p1fg}{HTML}{185B9D}
\definecolor{p2bg}{HTML}{E7F6F5}
\definecolor{p2fg}{HTML}{167A78}
\definecolor{p3bg}{HTML}{FFF0E3}
\definecolor{p3fg}{HTML}{B85A20}
\definecolor{p4bg}{HTML}{E8F6EC}
\definecolor{p4fg}{HTML}{11724F}

\providecommand{\learnstar}{\textcolor{learningorange}{\raisebox{-0.14ex}{{\fontsize{8.1}{8.1}\selectfont\ding{72}}}}}
\providecommand{\datastar}{\textcolor{datagreen}{\raisebox{-0.14ex}{{\fontsize{8.1}{8.1}\selectfont\ding{72}}}}}
\providecommand{\dimstar}{\textcolor{inactivegray}{\raisebox{-0.14ex}{{\fontsize{8.1}{8.1}\selectfont\ding{72}}}}}

\providecommand{\paradigmbadge}[3]{%
  \tikz[baseline=-0.55ex]{%
    \node[
      draw=#1,
      fill=#2,
      text=#1,
      rounded corners=2.0pt,
      line width=0.30pt,
      inner xsep=1.7pt,
      inner ysep=0.35pt,
      font=\bfseries\fontsize{6.4}{6.4}\selectfont
    ] {#3};%
  }%
}
\providecommand{\paradigmtag}[1]{%
  \ifcase#1\relax
  \or \paradigmbadge{p1fg}{p1bg}{PI}%
  \or \paradigmbadge{p2fg}{p2bg}{PII}%
  \or \paradigmbadge{p3fg}{p3bg}{PIII}%
  \or \paradigmbadge{p4fg}{p4bg}{PIV}%
  \fi
}

\begin{table*}[!t]
\centering
\begingroup
\renewcommand{\arraystretch}{0.80}
\setlength{\aboverulesep}{0.15ex}
\setlength{\belowrulesep}{0.15ex}
\setlength{\tabcolsep}{2.0pt}
\setlength{\arrayrulewidth}{0.35pt}
\arrayrulecolor{tablegrid}

\fontsize{6.8}{6.6}\selectfont
\caption{\textbf{Taxonomic Overview of Video Understanding Agent Methods.} Methods are ordered by publication year. Additional methods are reported in Table~\ref{tab:method_master_appendix}.}
\label{tab:method_master_main}
\begin{tabularx}{\textwidth}{
>{\raggedright\arraybackslash}p{3.65cm}|
>{\centering\arraybackslash}p{0.60cm}
>{\raggedright\arraybackslash}p{2.10cm}
>{\centering\arraybackslash}p{1.10cm}|
*{3}{>{\centering\arraybackslash}X}|
*{3}{>{\centering\arraybackslash}X}
}
\toprule
\multicolumn{4}{>{\columncolor{metabluelight}}c|}{\textbf{\textcolor{metablue}{Method Taxonomy}}} &
\multicolumn{3}{>{\columncolor{learningorangelight}}c|}{\textbf{\textcolor{learningorange}{Learning Paradigms}}} &
\multicolumn{3}{>{\columncolor{datagreenlight}}c}{\textbf{\textcolor{datagreen}{Data and Supervision}}} \\
\cmidrule(lr){1-4}\cmidrule(lr){5-7}\cmidrule(lr){8-10}
\cellcolor{metabluelight}\textbf{\textcolor{metablue}{Method}} &
\cellcolor{metabluelight}\textbf{\textcolor{metablue}{Year}} &
\cellcolor{metabluelight}\textbf{\textcolor{metablue}{\shortstack{Primary\\Challenge}}} &
\cellcolor{metabluelight}\textbf{\textcolor{metablue}{Paradigm}} &
\cellcolor{learningorangelight}\textbf{\textcolor{learningorange}{Training-Free}} &
\cellcolor{learningorangelight}\textbf{\textcolor{learningorange}{SFT}} &
\cellcolor{learningorangelight}\textbf{\textcolor{learningorange}{RL}} &
\cellcolor{datagreenlight}\textbf{\textcolor{datagreen}{Trajectory}} &
\cellcolor{datagreenlight}\textbf{\textcolor{datagreen}{Grounding}} &
\cellcolor{datagreenlight}\textbf{\textcolor{datagreen}{Reward}} \\
\midrule
AVT~\cite{yang2024agent} & 2024 & Temporal Causality & \paradigmtag{2} & \learnstar & \dimstar & \dimstar & \dimstar & \dimstar & \dimstar \\
\rowcolor{methodrowblue} DoraemonGPT~\cite{yang2024doraemongpt} & 2024 & Temporal Causality & \paradigmtag{3} & \learnstar & \dimstar & \dimstar & \dimstar & \dimstar & \dimstar \\
IXC2.5-OL~\cite{zhang2024internlm} & 2024 & Multimodal Ambiguity & \paradigmtag{4} & \learnstar & \dimstar & \dimstar & \dimstar & \dimstar & \dimstar \\
\rowcolor{methodrowblue} OmAgent~\cite{zhang2024omagent} & 2024 & Context Bottleneck & \paradigmtag{3} & \learnstar & \dimstar & \dimstar & \dimstar & \dimstar & \dimstar \\
SALOVA~\cite{kim2025salova} & 2024 & Context Bottleneck & \paradigmtag{1} & \dimstar & \learnstar & \dimstar & \dimstar & \dimstar & \dimstar \\
\rowcolor{methodrowblue} SlowFocus~\cite{nie2024slowfocus} & 2024 & Context Bottleneck & \paradigmtag{2} & \dimstar & \learnstar & \dimstar & \dimstar & \datastar & \dimstar \\
Video-of-Thought~\cite{pmlr-v235-fei24a} & 2024 & Temporal Causality & \paradigmtag{3} & \learnstar & \learnstar & \dimstar & \dimstar & \datastar & \dimstar \\
\rowcolor{methodrowblue} VideoAgent (Fan et al.)~\cite{fan2024videoagent} & 2024 & Context Bottleneck & \paradigmtag{3} & \learnstar & \dimstar & \dimstar & \dimstar & \dimstar & \dimstar \\
VideoAgent (Wang et al.)~\cite{wang2024videoagent} & 2024 & Evidence Sparsity & \paradigmtag{2} & \learnstar & \dimstar & \dimstar & \dimstar & \dimstar & \dimstar \\
\rowcolor{methodrowblue} VideoStreaming~\cite{qian2024streaming} & 2024 & Context Bottleneck & \paradigmtag{4} & \dimstar & \learnstar & \dimstar & \dimstar & \datastar & \dimstar \\
\specialrule{0.25pt}{0.25ex}{0.25ex}
AdaVideoRAG~\cite{zhang2026adavideorag} & 2025 & Context Bottleneck & \paradigmtag{3} & \learnstar & \dimstar & \dimstar & \dimstar & \dimstar & \dimstar \\
\rowcolor{methodrowblue} AoTD~\cite{shi2025enhancing} & 2025 & Evidence Sparsity & \paradigmtag{2} & \dimstar & \learnstar & \dimstar & \datastar & \dimstar & \dimstar \\
AVI~\cite{gao2025agentic} & 2025 & Context Bottleneck & \paradigmtag{3} & \learnstar & \dimstar & \dimstar & \dimstar & \dimstar & \dimstar \\
\rowcolor{methodrowblue} AViLA~\cite{zhang2025avila} & 2025 & Temporal Causality & \paradigmtag{4} & \learnstar & \dimstar & \dimstar & \dimstar & \dimstar & \dimstar \\
Commonsense Video QA~\cite{liu2025commonsense} & 2025 & Evidence Sparsity & \paradigmtag{2} & \learnstar & \dimstar & \dimstar & \dimstar & \dimstar & \dimstar \\
\rowcolor{methodrowblue} DrVideo~\cite{ma2025drvideo} & 2025 & Context Bottleneck & \paradigmtag{3} & \learnstar & \dimstar & \dimstar & \dimstar & \dimstar & \dimstar \\
DVD~\cite{zhang2026deep} & 2025 & Evidence Sparsity & \paradigmtag{3} & \learnstar & \dimstar & \dimstar & \dimstar & \dimstar & \dimstar \\
\rowcolor{methodrowblue} EgoAgent~\cite{chen2025egoagent} & 2025 & Temporal Causality & \paradigmtag{4} & \dimstar & \learnstar & \dimstar & \datastar & \dimstar & \dimstar \\
Embodied VideoAgent~\cite{fan2025embodied} & 2025 & Temporal Causality & \paradigmtag{4} & \learnstar & \dimstar & \dimstar & \dimstar & \dimstar & \dimstar \\
\rowcolor{methodrowblue} Eyes Wide Open~\cite{yu2026eyes} & 2025 & Temporal Causality & \paradigmtag{4} & \dimstar & \learnstar & \dimstar & \datastar & \dimstar & \dimstar \\
\rowcolor{methodrowblue} Flash-VStream~\cite{zhang2025flash} & 2025 & Context Bottleneck & \paradigmtag{4} & \dimstar & \learnstar & \dimstar & \dimstar & \dimstar & \dimstar \\
FrameThinker~\cite{he2025framethinker} & 2025 & Evidence Sparsity & \paradigmtag{1} & \dimstar & \learnstar & \learnstar & \datastar & \dimstar & \datastar \\
GraphVideoAgent~\cite{chu2025graphvideoagent} & 2025 & Temporal Causality & \paradigmtag{3} & \learnstar & \dimstar & \dimstar & \dimstar & \dimstar & \dimstar \\
\rowcolor{methodrowblue} LION-FS~\cite{li2025lion} & 2025 & Temporal Causality & \paradigmtag{4} & \dimstar & \learnstar & \dimstar & \datastar & \dimstar & \dimstar \\
LongVT~\cite{yang2026longvt} & 2025 & Evidence Sparsity & \paradigmtag{2} & \dimstar & \learnstar & \learnstar & \datastar & \dimstar & \datastar \\
\rowcolor{methodrowblue} LVAgent~\cite{lvagent2025} & 2025 & Context Bottleneck & \paradigmtag{1} & \learnstar & \dimstar & \dimstar & \dimstar & \dimstar & \dimstar \\
M3-Agent~\cite{long2025seeing} & 2025 & Context Bottleneck & \paradigmtag{3} & \dimstar & \learnstar & \learnstar & \dimstar & \datastar & \datastar \\
\rowcolor{methodrowblue} MAGNET~\cite{chowdhury2026magnet} & 2025 & Multimodal Ambiguity & \paradigmtag{3} & \learnstar & \dimstar & \dimstar & \dimstar & \dimstar & \dimstar \\
MAViD~\cite{chen2025multimodal} & 2025 & Multimodal Ambiguity & \paradigmtag{2} & \dimstar & \learnstar & \dimstar & \dimstar & \dimstar & \dimstar \\
\rowcolor{methodrowblue} Mobile-Agent-V~\cite{wang2025mobile} & 2025 & Temporal Causality & \paradigmtag{2} & \dimstar & \learnstar & \dimstar & \datastar & \dimstar & \dimstar \\
MONDAY~\cite{jang2025scalable} & 2025 & Temporal Causality & \paradigmtag{2} & \dimstar & \learnstar & \dimstar & \datastar & \dimstar & \dimstar \\
\rowcolor{methodrowblue} Mr. Video~\cite{pang2025mr} & 2025 & Context Bottleneck & \paradigmtag{3} & \learnstar & \dimstar & \dimstar & \dimstar & \dimstar & \dimstar \\
MSR-ViR~\cite{song2025modularized} & 2025 & Evidence Sparsity & \paradigmtag{2} & \dimstar & \learnstar & \learnstar & \dimstar & \datastar & \datastar \\
\rowcolor{methodrowblue} PANDA~\cite{yang2026panda} & 2025 & Temporal Causality & \paradigmtag{4} & \learnstar & \dimstar & \dimstar & \dimstar & \dimstar & \dimstar \\
ProVideLLM~\cite{chatterjee2025memory} & 2025 & Context Bottleneck & \paradigmtag{4} & \dimstar & \learnstar & \dimstar & \dimstar & \dimstar & \dimstar \\
\rowcolor{methodrowblue} ReAgent-V~\cite{reagentv2025} & 2025 & Evidence Sparsity & \paradigmtag{1} & \dimstar & \dimstar & \learnstar & \dimstar & \dimstar & \datastar \\
ReKV~\cite{di2025streaming} & 2025 & Context Bottleneck & \paradigmtag{4} & \learnstar & \dimstar & \dimstar & \dimstar & \dimstar & \dimstar \\
\rowcolor{methodrowblue} ReWatch-R1~\cite{zhang2025rewatch} & 2025 & Temporal Causality & \paradigmtag{2} & \dimstar & \learnstar & \learnstar & \datastar & \datastar & \datastar \\
RoomTour3D~\cite{han2025roomtour3d} & 2025 & Temporal Causality & \paradigmtag{2} & \dimstar & \learnstar & \dimstar & \datastar & \dimstar & \dimstar \\
\rowcolor{methodrowblue} SciEducator~\cite{xu2026scieducator} & 2025 & Multimodal Ambiguity & \paradigmtag{3} & \learnstar & \dimstar & \dimstar & \dimstar & \dimstar & \dimstar \\
SEASON~\cite{wu2026season} & 2025 & Temporal Causality & \paradigmtag{2} & \learnstar & \dimstar & \dimstar & \dimstar & \dimstar & \dimstar \\
\rowcolor{methodrowblue} Select Less, Reason More~\cite{li2026select} & 2025 & Evidence Sparsity & \paradigmtag{1} & \dimstar & \dimstar & \learnstar & \dimstar & \datastar & \datastar \\
StreamBridge~\cite{wang2026streambridge} & 2025 & Temporal Causality & \paradigmtag{4} & \dimstar & \learnstar & \dimstar & \datastar & \dimstar & \dimstar \\
\rowcolor{methodrowblue} StreamChat~\cite{xiong2025streaming} & 2025 & Context Bottleneck & \paradigmtag{4} & \learnstar & \dimstar & \dimstar & \dimstar & \dimstar & \dimstar \\
Thinking with Videos~\cite{zhang2026thinking} & 2025 & Evidence Sparsity & \paradigmtag{2} & \dimstar & \learnstar & \learnstar & \datastar & \dimstar & \datastar \\
\rowcolor{methodrowblue} Time-R1~\cite{wang2026time} & 2025 & Temporal Causality & \paradigmtag{2} & \dimstar & \dimstar & \learnstar & \dimstar & \dimstar & \datastar \\
\bottomrule
\end{tabularx}

\vspace{2pt}
\begin{minipage}{0.995\textwidth}
\fontsize{6.2}{7.0}\selectfont
\textit{Challenge notation:}\quad Primary Challenge reports the video-specific bottleneck most directly addressed by each method from the survey's analytical perspective.\par
\vspace{1pt}
\textit{Paradigm notation:}\quad
\paradigmtag{1}\,= Video as a Bag of Frames;\quad
\paradigmtag{2}\,= Video as a Sequence of Frames;\quad
\paradigmtag{3}\,= Video as a Graph of Entities;\quad
\paradigmtag{4}\,= Video as an Evolving World State.\par
\vspace{1pt}
\textit{Learning/Supervision notation:}\quad
SFT = Supervised Fine-Tuning;\quad
RL = Reinforcement Learning;\quad
Trajectory = Trajectory Supervision;\quad
Grounding = Grounding Supervision;\quad
Reward = Reward Signals.
\end{minipage}
\endgroup
\end{table*}

%
\definecolor{methodrowblue}{HTML}{F7F9FB}
\definecolor{tablegrid}{HTML}{B8C2C9}
\definecolor{metablue}{HTML}{185B9D}
\definecolor{metabluelight}{HTML}{EEF5FA}
\definecolor{learningorange}{HTML}{C65D1E}
\definecolor{learningorangelight}{HTML}{FDF3EA}
\definecolor{datagreen}{HTML}{11724F}
\definecolor{datagreenlight}{HTML}{EDF7F1}
\definecolor{inactivegray}{HTML}{D7DCE1}

\definecolor{p1bg}{HTML}{E5F1FC}
\definecolor{p1fg}{HTML}{185B9D}
\definecolor{p2bg}{HTML}{E7F6F5}
\definecolor{p2fg}{HTML}{167A78}
\definecolor{p3bg}{HTML}{FFF0E3}
\definecolor{p3fg}{HTML}{B85A20}
\definecolor{p4bg}{HTML}{E8F6EC}
\definecolor{p4fg}{HTML}{11724F}

\providecommand{\learnstar}{\textcolor{learningorange}{\raisebox{-0.14ex}{{\fontsize{8.1}{8.1}\selectfont\ding{72}}}}}
\providecommand{\datastar}{\textcolor{datagreen}{\raisebox{-0.14ex}{{\fontsize{8.1}{8.1}\selectfont\ding{72}}}}}
\providecommand{\dimstar}{\textcolor{inactivegray}{\raisebox{-0.14ex}{{\fontsize{8.1}{8.1}\selectfont\ding{72}}}}}

\providecommand{\paradigmbadge}[3]{%
  \tikz[baseline=-0.55ex]{%
    \node[
      draw=#1,
      fill=#2,
      text=#1,
      rounded corners=2.0pt,
      line width=0.30pt,
      inner xsep=1.7pt,
      inner ysep=0.35pt,
      font=\bfseries\fontsize{6.4}{6.4}\selectfont
    ] {#3};%
  }%
}
\providecommand{\paradigmtag}[1]{%
  \ifcase#1\relax
  \or \paradigmbadge{p1fg}{p1bg}{PI}%
  \or \paradigmbadge{p2fg}{p2bg}{PII}%
  \or \paradigmbadge{p3fg}{p3bg}{PIII}%
  \or \paradigmbadge{p4fg}{p4bg}{PIV}%
  \fi
}

\begin{table*}[!t]
\centering
\begingroup
\renewcommand{\arraystretch}{0.80}
\setlength{\aboverulesep}{0.15ex}
\setlength{\belowrulesep}{0.15ex}
\setlength{\tabcolsep}{2.0pt}
\setlength{\arrayrulewidth}{0.35pt}
\arrayrulecolor{tablegrid}

\fontsize{6.8}{6.6}\selectfont
\caption{\textbf{Taxonomic Overview of Video Understanding Agent Methods (continued).} Additional surveyed methods, ordered by publication year.}
\label{tab:method_master_appendix}
\begin{tabularx}{0.98\textwidth}{
>{\raggedright\arraybackslash}p{3.65cm}|
>{\centering\arraybackslash}p{0.60cm}
>{\raggedright\arraybackslash}p{2.10cm}
>{\centering\arraybackslash}p{1.10cm}|
*{3}{>{\centering\arraybackslash}X}|
*{3}{>{\centering\arraybackslash}X}
}
\toprule
\multicolumn{4}{>{\columncolor{metabluelight}}c|}{\textbf{\textcolor{metablue}{Method Taxonomy}}} &
\multicolumn{3}{>{\columncolor{learningorangelight}}c|}{\textbf{\textcolor{learningorange}{Learning Paradigms}}} &
\multicolumn{3}{>{\columncolor{datagreenlight}}c}{\textbf{\textcolor{datagreen}{Data and Supervision}}} \\
\cmidrule(lr){1-4}\cmidrule(lr){5-7}\cmidrule(lr){8-10}
\cellcolor{metabluelight}\textbf{\textcolor{metablue}{Method}} &
\cellcolor{metabluelight}\textbf{\textcolor{metablue}{Year}} &
\cellcolor{metabluelight}\textbf{\textcolor{metablue}{\shortstack{Primary\\Challenge}}} &
\cellcolor{metabluelight}\textbf{\textcolor{metablue}{Paradigm}} &
\cellcolor{learningorangelight}\textbf{\textcolor{learningorange}{Training-Free}} &
\cellcolor{learningorangelight}\textbf{\textcolor{learningorange}{SFT}} &
\cellcolor{learningorangelight}\textbf{\textcolor{learningorange}{RL}} &
\cellcolor{datagreenlight}\textbf{\textcolor{datagreen}{Trajectory}} &
\cellcolor{datagreenlight}\textbf{\textcolor{datagreen}{Grounding}} &
\cellcolor{datagreenlight}\textbf{\textcolor{datagreen}{Reward}} \\
\midrule
VCA~\cite{yang2025vca} & 2025 & Evidence Sparsity & \paradigmtag{2} & \learnstar & \dimstar & \dimstar & \dimstar & \dimstar & \dimstar \\
\rowcolor{methodrowblue} V-Agent~\cite{park2025v} & 2025 & Multimodal Ambiguity & \paradigmtag{1} & \dimstar & \learnstar & \dimstar & \dimstar & \dimstar & \dimstar \\
Video-R1~\cite{feng2026video} & 2025 & Temporal Causality & \paradigmtag{2} & \dimstar & \learnstar & \learnstar & \datastar & \dimstar & \datastar \\
\rowcolor{methodrowblue} Video-RAG~\cite{luo2026video} & 2025 & Context Bottleneck & \paradigmtag{3} & \learnstar & \dimstar & \dimstar & \dimstar & \dimstar & \dimstar \\
VideoAgent2~\cite{zhi2025videoagent2} & 2025 & Evidence Sparsity & \paradigmtag{2} & \learnstar & \dimstar & \dimstar & \dimstar & \dimstar & \dimstar \\
\rowcolor{methodrowblue} VideoAgentTrek~\cite{lu2025videoagenttrek} & 2025 & Temporal Causality & \paradigmtag{2} & \dimstar & \learnstar & \dimstar & \datastar & \dimstar & \dimstar \\
VideoExplorer~\cite{yuan2025videoexplorer} & 2025 & Evidence Sparsity & \paradigmtag{2} & \dimstar & \learnstar & \learnstar & \datastar & \dimstar & \datastar \\
\rowcolor{methodrowblue} VideoLLaMB~\cite{wang2025videollamb} & 2025 & Context Bottleneck & \paradigmtag{3} & \learnstar & \dimstar & \dimstar & \dimstar & \dimstar & \dimstar \\
VideoLucy~\cite{zuo2026videolucy} & 2025 & Context Bottleneck & \paradigmtag{3} & \learnstar & \dimstar & \dimstar & \dimstar & \dimstar & \dimstar \\
\rowcolor{methodrowblue} ViSpeak~\cite{fu2025vispeak} & 2025 & Multimodal Ambiguity & \paradigmtag{4} & \dimstar & \learnstar & \dimstar & \datastar & \dimstar & \dimstar \\
VITED~\cite{lu2025vited} & 2025 & Temporal Causality & \paradigmtag{2} & \dimstar & \learnstar & \dimstar & \datastar & \dimstar & \dimstar \\
\rowcolor{methodrowblue} VTimeCoT~\cite{zhang2025vtimecot} & 2025 & Temporal Causality & \paradigmtag{2} & \dimstar & \learnstar & \dimstar & \datastar & \dimstar & \dimstar \\
When Thinking Drifts~\cite{luo2026thinking} & 2025 & Temporal Causality & \paradigmtag{2} & \dimstar & \dimstar & \learnstar & \dimstar & \dimstar & \datastar \\
\specialrule{0.25pt}{0.25ex}{0.25ex}
\rowcolor{methodrowblue} A4VL~\cite{a4vl2026} & 2026 & Evidence Sparsity & \paradigmtag{1} & \learnstar & \dimstar & \dimstar & \dimstar & \dimstar & \dimstar \\
AgenticVS~\cite{guo2026agentic} & 2026 & Temporal Causality & \paradigmtag{2} & \learnstar & \learnstar & \dimstar & \datastar & \dimstar & \dimstar \\
\rowcolor{methodrowblue} APPO~\cite{du2026appo} & 2026 & Evidence Sparsity & \paradigmtag{1} & \dimstar & \dimstar & \learnstar & \dimstar & \dimstar & \datastar \\
EGAgent~\cite{rege2026agentic} & 2026 & Context Bottleneck & \paradigmtag{3} & \learnstar & \dimstar & \dimstar & \dimstar & \dimstar & \dimstar \\
\rowcolor{methodrowblue} EVA~\cite{zhang2026eva} & 2026 & Evidence Sparsity & \paradigmtag{1} & \dimstar & \learnstar & \learnstar & \dimstar & \dimstar & \datastar \\
G2F-RAG~\cite{yang2026graph} & 2026 & Multimodal Ambiguity & \paradigmtag{3} & \learnstar & \dimstar & \dimstar & \dimstar & \dimstar & \dimstar \\
\rowcolor{methodrowblue} HAVEN~\cite{yin2026hierarchical} & 2026 & Context Bottleneck & \paradigmtag{3} & \learnstar & \dimstar & \dimstar & \dimstar & \dimstar & \dimstar \\
LensWalk~\cite{li2026lenswalk} & 2026 & Evidence Sparsity & \paradigmtag{2} & \learnstar & \dimstar & \dimstar & \dimstar & \dimstar & \dimstar \\
\rowcolor{methodrowblue} LongVideo-R1~\cite{qiu2026longvideo} & 2026 & Evidence Sparsity & \paradigmtag{2} & \dimstar & \learnstar & \learnstar & \datastar & \dimstar & \datastar \\
OmniAgent~\cite{xing2026native} & 2026 & Context Bottleneck & \paradigmtag{3} & \dimstar & \learnstar & \learnstar & \datastar & \dimstar & \datastar \\
\rowcolor{methodrowblue} Proact-VL~\cite{yan2026proact} & 2026 & Temporal Causality & \paradigmtag{4} & \dimstar & \learnstar & \dimstar & \datastar & \datastar & \dimstar \\
R3-Streaming~\cite{liu2026efficient} & 2026 & Context Bottleneck & \paradigmtag{4} & \dimstar & \dimstar & \learnstar & \dimstar & \dimstar & \datastar \\
\rowcolor{methodrowblue} ThinkStream~\cite{liu2026thinking} & 2026 & Temporal Causality & \paradigmtag{4} & \dimstar & \dimstar & \learnstar & \dimstar & \dimstar & \datastar \\
Refer-Agent~\cite{jiang2026referagent} & 2026 & Evidence Sparsity & \paradigmtag{1} & \learnstar & \dimstar & \dimstar & \dimstar & \datastar & \dimstar \\
\rowcolor{methodrowblue} ReViSe~\cite{xu2026towards} & 2026 & Evidence Sparsity & \paradigmtag{1} & \learnstar & \dimstar & \learnstar & \dimstar & \dimstar & \datastar \\
SAGE~\cite{jain2026sage} & 2026 & Evidence Sparsity & \paradigmtag{2} & \dimstar & \learnstar & \learnstar & \datastar & \dimstar & \datastar \\
\rowcolor{methodrowblue} StreamAgent~\cite{yang2025streamagent} & 2026 & Temporal Causality & \paradigmtag{4} & \dimstar & \learnstar & \dimstar & \datastar & \dimstar & \dimstar \\
Streaming Harness~\cite{yao2026harnessing} & 2026 & Temporal Causality & \paradigmtag{4} & \dimstar & \learnstar & \dimstar & \datastar & \dimstar & \dimstar \\
\rowcolor{methodrowblue} StreamMeCo~\cite{wang2026streammeco} & 2026 & Context Bottleneck & \paradigmtag{3} & \learnstar & \dimstar & \dimstar & \dimstar & \dimstar & \dimstar \\
StreamRAG~\cite{xie2026streamrag} & 2026 & Context Bottleneck & \paradigmtag{4} & \learnstar & \dimstar & \dimstar & \dimstar & \dimstar & \dimstar \\
\rowcolor{methodrowblue} StreamReady~\cite{azad2026streamready} & 2026 & Temporal Causality & \paradigmtag{4} & \dimstar & \learnstar & \dimstar & \dimstar & \datastar & \dimstar \\
SVAgent~\cite{yang2026svagent} & 2026 & Temporal Causality & \paradigmtag{3} & \learnstar & \dimstar & \dimstar & \dimstar & \dimstar & \dimstar \\
\rowcolor{methodrowblue} Symphony~\cite{yan2026symphony} & 2026 & Context Bottleneck & \paradigmtag{2} & \learnstar & \dimstar & \dimstar & \dimstar & \dimstar & \dimstar \\
Takusen~\cite{tang2026asynchronous} & 2026 & Temporal Causality & \paradigmtag{4} & \dimstar & \learnstar & \dimstar & \datastar & \dimstar & \dimstar \\
\rowcolor{methodrowblue}
Thinking-QwenVL~\cite{zhang2026progressive} & 2026 & Temporal Causality & \paradigmtag{4} & \dimstar & \learnstar & \dimstar & \dimstar & \datastar & \dimstar \\
Video-MTR~\cite{xie2025video} & 2026 & Evidence Sparsity & \paradigmtag{2} & \dimstar & \dimstar & \learnstar & \dimstar & \dimstar & \datastar \\
\rowcolor{methodrowblue} VideoARM~\cite{yin2026videoarm} & 2026 & Context Bottleneck & \paradigmtag{3} & \learnstar & \dimstar & \dimstar & \dimstar & \dimstar & \dimstar \\
VideoBrain~\cite{zou2026videobrain} & 2026 & Evidence Sparsity & \paradigmtag{1} & \dimstar & \learnstar & \learnstar & \datastar & \dimstar & \datastar \\
\rowcolor{methodrowblue} VideoChat-A1~\cite{wang2026videochat} & 2026 & Evidence Sparsity & \paradigmtag{2} & \learnstar & \dimstar & \dimstar & \dimstar & \dimstar & \dimstar \\
VideoChat-M1~\cite{chen2026videochat} & 2026 & Evidence Sparsity & \paradigmtag{2} & \dimstar & \dimstar & \learnstar & \dimstar & \dimstar & \datastar \\
\rowcolor{methodrowblue} VideoHV-Agent~\cite{wang2026think} & 2026 & Evidence Sparsity & \paradigmtag{2} & \learnstar & \dimstar & \dimstar & \dimstar & \dimstar & \dimstar \\
VideoMind~\cite{liu2025videomind} & 2026 & Temporal Causality & \paradigmtag{2} & \dimstar & \learnstar & \dimstar & \dimstar & \datastar & \dimstar \\
\rowcolor{methodrowblue} VideoSEAL~\cite{qiu2026videoseal} & 2026 & Evidence Sparsity & \paradigmtag{2} & \dimstar & \dimstar & \learnstar & \dimstar & \datastar & \datastar \\
VideoSeek~\cite{lin2026videoseek} & 2026 & Evidence Sparsity & \paradigmtag{2} & \learnstar & \dimstar & \dimstar & \dimstar & \dimstar & \dimstar \\
\rowcolor{methodrowblue} WorldMM~\cite{yeo2026worldmm} & 2026 & Context Bottleneck & \paradigmtag{3} & \learnstar & \dimstar & \dimstar & \dimstar & \dimstar & \dimstar \\
\bottomrule
\end{tabularx}

\vspace{2pt}
\begin{minipage}{0.995\textwidth}
\fontsize{6.2}{7.0}\selectfont
\textit{Challenge notation:}\quad Primary Challenge reports the video-specific bottleneck most directly addressed by each method from the survey's analytical perspective.\par
\vspace{1pt}
\textit{Paradigm notation:}\quad
\paradigmtag{1}\,= Video as a Bag of Frames;\quad
\paradigmtag{2}\,= Video as a Sequence of Frames;\quad
\paradigmtag{3}\,= Video as a Graph of Entities;\quad
\paradigmtag{4}\,= Video as an Evolving World State.\par
\vspace{1pt}
\textit{Learning/Supervision notation:}\quad
SFT = Supervised Fine-Tuning;\quad
RL = Reinforcement Learning;\quad
Trajectory = Trajectory Supervision;\quad
Grounding = Grounding Supervision;\quad
Reward = Reward Signals.
\end{minipage}
\endgroup
\end{table*}

\section{Learning Paradigms for Video Agents}
The preceding sections describe why agentic control is needed and what video states agents operate on. We now examine how agent behavior is specified or learned. We group existing methods by their dominant mechanism: Training-Free and Inference-Time Control; Supervised Fine-Tuning (SFT) and Imitation Learning, and Reinforcement Learning (RL). These categories are not mutually exclusive, since multi-stage systems may combine supervised initialization with reinforcement learning.

Tables~\ref{tab:method_master_main} and~\ref{tab:method_master_appendix}
summarize the surveyed methods across their primary challenge,
state space paradigm, learning paradigm, and supervision signals.

\subsection{Training-Free and Inference-Time Control}
Training-free video agents do not optimize a dedicated agent policy.
Instead, they assemble behavior at inference time through prompts, tool interfaces, retrieval procedures, memory rules, verification mechanisms, and stopping or waiting criteria.

The first group follows prompted tool-use and search. VideoAgent~\cite{wang2024videoagent}, OmAgent~\cite{zhang2024omagent}, A4VL~\cite{a4vl2026}, VCA~\cite{yang2025vca}, LensWalk~\cite{li2026lenswalk}, VideoSeek~\cite{lin2026videoseek}, Deep Video Discovery~\cite{zhang2026deep}, and Agentic Video Intelligence~\cite{gao2025agentic} ask the model to decompose the query, inspect temporal regions or memory entries, call video tools, update intermediate evidence, and synthesize an answer. Their agentic behavior is not learned from trajectories or rewards, but specified through prompts, retrieval modules, external tools, and video states.

Role decomposition can also be implemented in a training-free manner. Symphony~\cite{yan2026symphony} separates planning, grounding, subtitle reasoning, visual perception, and reflection into specialized agents. SVAgent~\cite{yang2026svagent} assigns visual, textual, and meta-level decision roles for storyline-guided reasoning, while SciEducator~\cite{xu2026scieducator} organizes scientific video understanding as a Deming-cycle collaboration among agents. These systems show that multi-agent structure can support modularity and verification even when the roles themselves are prompt-defined rather than learned.

The second group emphasizes inference-time correction and reliability control. DrVideo~\cite{ma2025drvideo} allows the agent to judge whether the current video document contains sufficient evidence and retrieve missing visual details when needed. VideoAgent2~\cite{zhi2025videoagent2} uses uncertainty-aware reasoning to adjust tool calls. VideoHV-Agent~\cite{wang2026think} turns candidate answers into hypotheses to be checked against localized evidence. VideoLucy~\cite{zuo2026videolucy} performs confidence-driven memory backtracking, while SEASON~\cite{wu2026season} uses self-diagnostic contrastive decoding to reduce temporal hallucination. AViLA~\cite{zhang2025avila} further extends this paradigm to streaming interaction, where the agent decides whether to answer, retrieve past evidence, or wait for future evidence.

A practical advantage of this paradigm is accessibility: it can exploit strong foundation models without collecting process-level labels. Its weakness is policy fragility. Inference-time agents may oversearch, stop too early, repeat tool calls, trust noisy tools, or follow fluent but weakly grounded plans.

\subsection{Supervised Fine-Tuning (SFT) and Imitation Learning}
Supervised learning trains video-agent behavior at different levels of granularity. Answer-level SFT learns to generate a response from a video and an instruction, but provides limited guidance on how the agent should obtain that response. Imitation learning instead supervises demonstrated decisions or trajectories. In current video agents, it is commonly implemented as SFT on teacher-generated traces; the distinction therefore lies mainly in the supervision target rather than the optimization algorithm.

Trajectory imitation often initializes an agent before preference or reinforcement learning. VideoExplorer~\cite{yuan2025videoexplorer} explicitly performs structured imitation by training its planner and temporal grounder on successful reasoning trajectories. LongVideo-R1~\cite{qiu2026longvideo} learns from chain-of-thought-with-tool demonstrations that specify when to inspect a video segment, continue reasoning, or terminate. SAGE~\cite{jain2026sage} similarly constructs tool-use trajectories with a stronger model and converts them into state--action pairs for cold-start SFT. In these systems, imitation establishes an initial policy, while later optimization improves its effectiveness.

Other methods apply supervised learning to selected components rather than complete trajectories. VideoMind~\cite{liu2025videomind} separately trains planner, grounder, and verifier roles using planning labels, temporal localization objectives, and verification labels, with Chain-of-LoRA enabling role switching at inference. AoTD~\cite{shi2025enhancing} transfers verified reasoning traces produced by an external agent system into a single Video-LLM, while VITED~\cite{lu2025vited} trains models to generate automatically constructed temporal evidence chains. 

Supervised learning provides direct, dense supervision where labels or traces are available, but remains constrained by the coverage and quality of its demonstrations or labels. It can teach an agent to reproduce successful patterns without ensuring that it can adapt when unfamiliar states require a different strategy.

\subsection{Reinforcement Learning}
Reinforcement learning (RL) optimizes video agent decisions through outcome- or process-level rewards rather than only imitating demonstrated trajectories. Rewards may measure answer correctness, evidence grounding, action efficiency, response timing, or reasoning validity. This is particularly useful for learning decisions such as where to inspect, which evidence to retain, when to invoke a tool, and when sufficient evidence has been collected to answer.

One line of work applies RL to perception and evidence acquisition. SAGE~\cite{jain2026sage} trains agents to choose among direct answering, rapid skimming, and deeper video inspection. Thinking with Videos~\cite{zhang2026thinking} learns tool-augmented policies for acquiring evidence from long videos, while APPO~\cite{du2026appo} optimizes attention-guided selection of informative visual tokens. LongVideo-R1~\cite{qiu2026longvideo} learns navigation over hierarchical video structures. R3-Streaming~\cite{liu2026efficient} extends such control to streaming settings by optimizing memory forgetting, answer readiness, and fast--slow processing routes.

RL can also optimize coordination among multiple agents. ReAgent-V~\cite{reagentv2025} supports self-correction through a critic agent that provides feedback and generates follow-up questions for the target agent. VideoChat-M1~\cite{chen2026videochat} uses collaborative planning and multi-agent reinforcement learning to optimize how specialized agents propose, execute, and revise tool-use plans. Unlike systems whose coordination is defined primarily through prompts, these methods make coordination decisions part of the learned behavior.

Another line focuses on verifiable rewards for video reasoning and grounding without learning a broader perception-and-tool control policy. Time-R1~\cite{wang2026time} uses localization-oriented rewards for temporal grounding, while Video-R1~\cite{feng2026video} improves temporal reasoning through RL post-training. DeepVideo-R1~\cite{park2026deepvideo} introduces difficulty-aware regressive GRPO, and When Thinking Drifts~\cite{luo2026thinking} rewards reasoning trajectories that remain grounded in visual evidence.

RL suits video agents well because many decisions lack direct supervision. A central challenge is credit assignment: sparse rewards often cannot reveal whether success comes from a reliable policy or accidental shortcuts.

\section{Data and Supervision for Video Agents}
Learning richer agent behaviors can benefit from supervision beyond video-question-answer pairs. A useful training example may need to show which evidence the agent inspected, which tool it called, what memory it wrote, which temporal interval grounded the claim, and why the agent stopped, revised, or responded. This section separates three data-related aspects: trajectory supervision, grounding supervision, and preference and reward supervision as a supervision source.

\subsection{Trajectory Supervision}
Trajectory supervision records the step-by-step path an agent takes to reach an answer or action. A trace should record the agent's observations, decisions, state changes, and final response; richer traces also capture failed branches, revisions, stopping decisions, and cost. Such data is useful not only for imitation learning, but also for auditing whether a video agent followed a grounded decision path rather than merely producing a plausible final answer.

For video understanding, trajectory supervision is more complex than in text-only agents. The trace should identify the video units involved in each step, such as frames, clips, transcript spans, audio segments, object tracks, temporal segments, or memory entries, and record how these observations changed the next action. Agent-of-Thoughts Distillation~\cite{shi2025enhancing} distills tool-supported decomposition and reasoning traces into a video model. ReWatch-R1~\cite{zhang2025rewatch} bootstraps timestamped, multi-agent Reasoning and Acting (ReAct)-style traces for complex video reasoning. VITED~\cite{lu2025vited} distills temporal evidence chains, while VTimeCoT~\cite{zhang2025vtimecot} represents reasoning as an annotated visual temporal canvas. VideoExplorer~\cite{yuan2025videoexplorer} further organizes long-video reasoning as trajectories that connect planning, temporal grounding, and evidence reading.

Interactive video data adds another layer: the timestamp of each observation, query, interruption, and response may matter. StreamBridge~\cite{wang2026streambridge} constructs interleaved video-text instruction traces for streaming assistants, Takusen~\cite{tang2026asynchronous} supervises asynchronous dense captioning through event-start and event-end decisions, and ViSpeak~\cite{fu2025vispeak} includes visual instructions such as wake-up, reference, interruption, and termination.

\subsection{Grounding Supervision}
Grounding supervision links claims, answers, intermediate states, and actions to their supporting video evidence. It includes temporal intervals, spatial regions, entity identities, object tracks, audio cues, transcript spans, OCR evidence, and state changes. Temporal evidence alignment is especially important because a response may be semantically plausible yet supported by evidence from the wrong moment, while a reasoning trace may also drift away from the visual evidence actually observed~\cite{zhang2026progressive,luo2026thinking}.

Grounded reasoning systems point toward this requirement. VideoMind~\cite{liu2025videomind} grounds reasoning in temporal candidate moments before answer generation. VideoHV-Agent~\cite{wang2026think} turns candidate answers into hypotheses and verifies them against localized video evidence. Refer-Agent~\cite{jiang2026referagent} extends grounding to spatial evidence by checking object existence, identity consistency, and mask reliability. A4VL~\cite{a4vl2026} aligns query-specific perception clues with relevant video blocks, while MAGNET~\cite{chowdhury2026magnet} stresses audio-visual grounding across multi-video haystacks. In the long run, grounding supervision should include negative evidence as well as positive evidence: the agent should know not only which segments support an answer, but also which inspected segments fail to support competing hypotheses.

Interactive and causal settings further expand grounding from asking where the evidence is to asking when it became available. AViLA~\cite{zhang2025avila} highlights query-evidence asynchrony, StreamEQA~\cite{wang2026streameqa} attaches questions to timestamps, and Streaming-Eval~\cite{yao2026harnessing} evaluates punctual behavior. In such settings, a correct answer may still be invalid if it relies on future evidence unavailable at decision time.

\subsection{Preference and Reward Supervision}
Preference and reward supervision provides comparative or scalar signals for learning which agent behavior to prefer. Unlike trajectory supervision, which records the process, or grounding supervision, which anchors claims to evidence, this form of data evaluates the quality of answers, evidence choices, reasoning traces, tool calls, or complete rollouts. 

Recent work illustrates this supervision at different granularities. V-Agent~\cite{park2025v} fine-tunes a retrieval-oriented VLM with a small video preference dataset, providing preference signals for video search rather than full agent rollouts. DeepVideo-R1~\cite{park2026deepvideo} organizes reward supervision around sample difficulty through regressive GRPO, while Video-R1~\cite{feng2026video} introduces temporal-aware rewards to encourage reasoning grounded in video dynamics. ThinkStream~\cite{liu2026thinking} derives automatically verifiable rewards for incremental reasoning and watch-think-speak timing. R3-Streaming~\cite{liu2026efficient} further uses routing rewards that reflect response readiness, latency, memory pressure, and fast--slow model selection. 

The main challenge is reward validity. If rewards only reflect answer correctness, they may miss whether the agent searched efficiently, used reliable evidence, or avoided spurious reasoning.

\section{Benchmarks and Evaluation}

\begin{table*}[!t]
\centering

\caption{Comparison of benchmarks relevant to agentic video understanding.}
\label{tab:benchmark_summary}

\scriptsize
\color{TableText}

\setlength{\tabcolsep}{2.1pt}
\renewcommand{\arraystretch}{0.98}
\arrayrulecolor{TableGrid}

\begin{tabularx}{\textwidth}{
@{}
L{0.185\textwidth}
!{\color{TableGrid}\vrule width 0.35pt}
>{\centering\arraybackslash\color{FigOrangeDark}}p{0.040\textwidth}
C{0.060\textwidth}
!{\color{TableGrid}\vrule width 0.35pt}
C{0.060\textwidth}
L{0.105\textwidth}
L{0.105\textwidth}
!{\color{TableGrid}\vrule width 0.35pt}
X
!{\color{TableGrid}\vrule width 0.35pt}
C{0.030\textwidth}
@{}
}

\toprule[0.9pt]

\rowcolor{TableHead}
\textbf{\textcolor{TableNavy}{Benchmark}}
&
\textbf{\textcolor{TableNavy}{Year}}
&
\textbf{\textcolor{TableNavy}{Video Mode}}
&
\textbf{\textcolor{TableNavy}{Len. (s)}}
&
\textbf{\textcolor{TableNavy}{Video Scale}}
&
\textbf{\textcolor{TableNavy}{Evaluation Items}}
&
\textbf{\textcolor{TableNavy}{Evaluation Format}}
&
\textbf{\textcolor{TableNavy}{Link}}
\\

\midrule[0.65pt]

\rowcolor{FigBlueLight}
\multicolumn{8}{@{}l@{}}{%
  \hspace{3pt}
  \textbf{\textcolor{FigBlueDark}{Offline / Fixed-Access Benchmarks}}
}
\\[-1pt]

Charades-STA~\cite{gao2017tall}
& 2017
& \offlinebadge
& $\sim$30
& 1,334 videos
& 3,720
& Temporal grounding
& \benchlink{https://github.com/jiyanggao/TALL/tree/master}
\\

\rowcolor{FigBlueStripe}
MSRVTT-QA~\cite{10.1145/3123266.3123427}
& 2017
& \offlinebadge
& 15.2
& 2,990 videos
& 72,821
& Open-ended QA
& \benchlink{https://github.com/xudejing/video-question-answering}
\\

YouCook2~\cite{zhou2018towards}
& 2017
& \offlinebadge
& 315.6
& 210 videos
& 210
& Procedure segmentation
& \benchlink{http://youcook2.eecs.umich.edu/}
\\

\rowcolor{FigBlueStripe}
ActivityNet-QA~\cite{yu2019activitynet}
& 2019
& \offlinebadge
& 180.0
& 800 videos
& 8,000
& Open-ended QA
& \benchlink{https://github.com/MILVLG/activitynet-qa}
\\

How2QA~\cite{li2020hero}
& 2020
& \offlinebadge
& 60.0
& 1,166 videos
& 2,852
& MCQ + moment localization
& \benchlink{https://github.com/linjieli222/HERO}
\\

\rowcolor{FigBlueStripe}
NExT-QA~\cite{xiao2021next}
& 2021
& \offlinebadge
& 44.0
& 1,000 videos
& 8,564
& MCQ + open-ended QA
& \benchlink{https://github.com/doc-doc/NExT-QA}
\\

STAR~\cite{wu2024star}
& 2021
& \offlinebadge
& 29.7
& 955 videos
& 7,377
& MCQ
& \benchlink{https://bobbywu.com/STAR/}
\\

\rowcolor{FigBlueStripe}
EgoSchema~\cite{mangalam2023egoschema}
& 2023
& \offlinebadge
& 180.0
& 5,063 clips
& 5,063
& MCQ
& \benchlink{https://github.com/egoschema/egoschema}
\\

MVBench~\cite{li2024mvbench}
& 2023
& \offlinebadge
& 16.0
& 3,641 videos
& 4,000
& MCQ
& \benchlink{https://github.com/OpenGVLab/Ask-Anything}
\\

\rowcolor{FigBlueStripe}
Ego-Exo4D~\cite{grauman2024ego}
& 2023
& \offlinebadge
& 156.0
& 1,121 test takes
& Task-dependent
& Multi-task video understanding benchmark
& \benchlink{https://github.com/facebookresearch/Ego4d}
\\

HiREST~\cite{zala2023hierarchical}
& 2023
& \offlinebadge
& 263
& 1,391 videos
& 546
& Video retrieval; moment retrieval \& segmentation; step captioning
& \benchlink{https://github.com/j-min/HiREST}
\\

\rowcolor{FigBlueStripe}
Perception Test~\cite{patraucean2023perception}
& 2023
& \offlinebadge
& 23.0
& 3,525 videos
& Task-dependent
& Tracking; action/sound localization; MCQ; grounded QA
& \benchlink{https://github.com/google-deepmind/perception_test}
\\

Video-Bench~\cite{ning2025video}
& 2023
& \offlinebadge
& 56.0
& 5,917 videos
& 17,036
& MCQ
& \benchlink{https://github.com/PKU-YuanGroup/Video-Bench}
\\

\rowcolor{FigBlueStripe}
LongVideoBench~\cite{wu2024longvideobench}
& 2024
& \offlinebadge
& 473.0
& 3,011 videos
& 5,341
& MCQ
& \benchlink{https://huggingface.co/datasets/longvideobench/LongVideoBench}
\\

LVBench~\cite{wang2025lvbench}
& 2024
& \offlinebadge
& 4,101.0
& 103 videos
& 1,549
& MCQ
& \benchlink{https://huggingface.co/datasets/zai-org/LVBench}
\\

\rowcolor{FigBlueStripe}
MLVU~\cite{zhou2025mlvu}
& 2024
& \offlinebadge
& 180.0--7,200
& 1,334 videos (dev)
& 2,593 (dev)
& MCQ + open-ended generation
& \benchlink{https://github.com/JUNJIE99/MLVU}
\\

TempCompass~\cite{liu2024tempcompass}
& 2024
& \offlinebadge
& 11.4
& 410 videos
& 7,540
& MCQ + binary QA + captioning
& \benchlink{https://github.com/llyx97/TempCompass}
\\

\rowcolor{FigBlueStripe}
Video-MME~\cite{fu2025video}
& 2024
& \offlinebadge
& 1,017.9
& 900 videos
& 2,700
& MCQ
& \benchlink{https://github.com/MME-Benchmarks/Video-MME}
\\

MMBench-Video~\cite{fang2024mmbench}
& 2024
& \offlinebadge
& 165.4
& 609 videos
& 1,998
& Open-ended QA
& \benchlink{https://github.com/open-compass/VLMEvalKit}
\\

\rowcolor{FigBlueStripe}
LongVALE~\cite{geng2025longvale}
& 2024
& \offlinebadge
& 235.0
& 1,171 videos
& 13,867
& Temporal grounding; dense video captioning \& segment captioning
& \benchlink{https://huggingface.co/datasets/ttgeng233/LongVALE}
\\

Video-MMMU~\cite{hu2026video}
& 2025
& \offlinebadge
& 506.2
& 300 videos
& 900
& MCQ
& \benchlink{https://huggingface.co/datasets/lmms-lab/VideoMMMU}
\\

\rowcolor{FigBlueStripe}
Video-TT~\cite{zhang2025towards}
& 2025
& \offlinebadge
& $<65$
& 1,000 videos
& 5,000
& Open-ended QA; MCQ; natural-adversarial robustness
& \benchlink{https://huggingface.co/datasets/lmms-lab/video-tt}
\\

Ego2Web~\cite{yu2026ego2web}
& 2026
& \offlinebadge
& 180.0
& 500 pairs
& 500
& Web-task completion
& \benchlink{https://huggingface.co/datasets/Shoubin/Ego2Web}
\\

\rowcolor{FigBlueStripe}
VideoDR~\cite{liu2026watching}
& 2026
& \offlinebadge
& --
& 500
& 500
& Open-web factoid QA; browser search
& \benchlink{https://huggingface.co/datasets/strike20023/VideoDR}
\\

\midrule[0.55pt]

\rowcolor{FigGreenLight}
\multicolumn{8}{@{}l@{}}{%
  \hspace{3pt}
  \textbf{\textcolor{FigGreenDark}{Online / Streaming Benchmarks}}
}
\\[-1pt]

StreamingBench~\cite{lin2026streamingbench}
& 2024
& \onlinebadge
& 3--1,440
& 900 videos
& 4,500
& Timestamped MCQ; sequential QA; proactive output
& \benchlink{https://huggingface.co/datasets/mjuicem/StreamingBench}
\\

\rowcolor{FigGreenStripe}
OVO-Bench~\cite{niu2025ovo}
& 2025
& \onlinebadge
& --
& 644 videos
& 3,100
& MCQ + open-ended QA
& \benchlink{https://huggingface.co/datasets/JoeLeelyf/OVO-Bench}
\\

OmniMMI~\cite{wang2025omnimmi}
& 2025
& \onlinebadge
& 324.3
& 1,121 videos
& 2,290
& Streaming QA; state grounding; multi-turn and proactive response
& \benchlink{https://huggingface.co/datasets/bigai-nlco/OmniMMI}
\\

\rowcolor{FigGreenStripe}
OmniPro~\cite{zhao2026omnipro}
& 2026
& \onlinebadge
& 189.0
& 1,262 videos
& 2,700
& Proactive streaming QA; response timing
& \benchlink{https://huggingface.co/datasets/RuixiangZhao/OmniPro}
\\
\bottomrule[0.9pt]

\end{tabularx}

\vspace{3pt}

\begin{minipage}{0.99\textwidth}
\footnotesize
\color{TableText}
\textit{Notes.}
Year denotes the first public release of each benchmark.
Video Mode describes access to video evidence rather than whether an
external website or tool is online.
Evaluation Items reports the number of evaluation instances.
For benchmarks with training and test splits, reported scale statistics
refer to the test set unless otherwise noted.
Duration values are averages unless a range or upper bound is shown.
MCQ = multiple-choice question; QA = question answering.
\end{minipage}

\end{table*}
Benchmarks differ along two complementary dimensions: their evaluation role and their video-access regime.\textbf{ Capability-oriented benchmarks} measure skills required by video agents, including temporal grounding, long-range evidence integration, multimodal reasoning, and procedural understanding.\textbf{ Agent-oriented benchmarks} additionally evaluate how systems control evidence access, response timing, tool use, or external actions. Because these dimensions are independent, Table~\ref{tab:benchmark_summary} groups benchmarks by video-access regime, while the discussion identifies their primary evaluation role.

Capability-oriented evaluation has expanded from short-video question answering (QA),
temporal localization, and procedure understanding to long-context and
diagnostic reasoning. Early benchmarks such as Charades-STA~\cite{gao2017tall}, ActivityNet-QA~\cite{yu2019activitynet},
YouCook2~\cite{zhou2018towards} and NExT-QA~\cite{xiao2021next} established temporal grounding, open-ended QA,
instructional-video understanding, and causal reasoning. More recent benchmarks, including EgoSchema~\cite{mangalam2023egoschema}, MVBench~\cite{li2024mvbench}, LongVideoBench~\cite{wu2024longvideobench}, LVBench~\cite{wang2025lvbench}, MLVU~\cite{zhou2025mlvu},
TempCompass~\cite{liu2024tempcompass}, and Video-MME~\cite{fu2025video}, extend evaluation to long-range egocentric
reasoning, diagnostic evaluation of temporal and multimodal capabilities, and increasingly long video contexts.
These benchmarks measure important capabilities required by video
agents, but typically provide the complete video before prediction.

Agent-oriented evaluation adds constraints on when evidence becomes
available and what the system must decide or do. StreamingBench
~\cite{lin2026streamingbench} bridges conventional and agent-oriented
evaluation through timestamped video prefixes, sequential QA, and
proactive-output simulation. OVO-Bench~\cite{niu2025ovo},
OmniMMI~\cite{wang2025omnimmi}, and
OmniPro~\cite{zhao2026omnipro} further require causal video access and
decisions about when to respond. Ego2Web~\cite{yu2026ego2web} and
VideoDR~\cite{liu2026watching} extend video-conditioned evaluation to
web execution and external information retrieval. These benchmarks
shift the question from only what a model can answer to when the
evidence is available and what the system should do next.

Across these benchmarks, evaluation varies along three axes:
the \emph{evidence horizon}, from short clips to distributed
long-video evidence; the \emph{access regime}, from complete-video
input to causal observation; and the \emph{action scope}, from answer
generation to proactive response and external execution. However,
most benchmarks still emphasize final-task success. More complete evaluation should assess not only final-task success, but also the quality, timing, cost, and reliability of the agent's evidence-use process.

\section{Open Challenges and Future Directions}

\subsection{Agentic-Native Temporal Modeling}

Existing video models encode time through temporal convolution, recurrent
states, space--time attention, or timestamp-aware tokens and memory
structures~\cite{tran2015learning,carreira2017quo,donahue2015long,
bertasius2021space,yang2023vid2seq,ren2024timechat,
huang2024vtimellm,song2024moviechat}. For video agents, temporal information can play a more active role when it is exposed as an operable state rather than encoded only as part of the input. The agent should be able to query what has been
observed, trace the evidence supporting its current beliefs, revise
uncertain states, and use this information to guide later verification,
response, or action.

Current video agents distribute these functions across search, memory,
verification, and streaming-control modules~\cite{wang2024videoagent,
li2026lenswalk,lin2026videoseek,ma2025drvideo,yang2026svagent,
zhi2025videoagent2,wang2026think,liu2025videomind,zhang2025avila,
liu2026thinking,lin2026streamingbench,wang2025omnimmi}. A key direction
is to unify them through a temporal-state interface that supports
explicit grounding, reading, updating, and stopping operations. Such an
interface should connect temporally anchored evidence and uncertainty to
decisions about whether to observe, retrieve, verify, wait, respond, or
act. Evaluation should test whether agents preserve correct temporal
beliefs, update them as evidence arrives, select the right moments, and
avoid premature or unsupported decisions.

\subsection{Video-Native Agent Modeling}

Many agentic video systems in our corpus use an LLM or MLLM as the central reasoner. Video content is often converted into textual summaries, retrieved evidence, structured memories, or multimodal tokens, after which most reasoning occurs in a language-dominated space~\cite{wang2024videoagent,
ma2025drvideo,zhang2024omagent,fan2024videoagent,
chu2025graphvideoagent,yang2026svagent}. Although effective for instruction following, tool use, and retrieval, this pipeline may lose fine-grained visual dynamics such as motion, state changes, interactions, and action consequences.

Video-native agent modeling instead asks whether perceptual backbones
can represent these dynamics directly and expose them to downstream
agent reasoning. Future agents may use representations that capture
motion, state transitions, predictive scene evolution, and
action-conditioned outcomes, rather than treating video models mainly
as feature providers for an LLM controller. Research on world models,
generative control, and video-to-action learning provides useful
building blocks for this direction~\cite{hassan2025gem,
zhao2025drivedreamer4d,ma2024hierarchical,chen2025egoagent,
lu2025videoagenttrek,jang2025scalable}. A central question is whether
such backbones improve the perception of state changes, action anticipation, and downstream task success, particularly in procedural, manipulation, and embodied settings.

\section{Conclusion}
Video understanding agents shift video modeling from fixed inference to adaptive evidence control. This survey organizes the field through video-specific challenges, state-space paradigms, learning mechanisms, supervision, and evaluation.  We highlight two promising directions: agentic-native video modeling for operable temporal states and video-native agent modeling for reasoning over visual dynamics. Evaluation should also extend beyond final accuracy to evidence use, response timing, tool decisions, cost, and interaction reliability.
\bibliographystyle{IEEEtran}
\bibliography{video_understanding_agent_references}

\end{document}